%% file: main.tex
\documentclass[10pt,twocolumn,letterpaper]{article}
\usepackage[table, dvipsnames]{xcolor}
\usepackage[pagenumbers]{cvpr} 

\input{preamble}

\definecolor{cvprblue}{rgb}{0.21,0.49,0.74}
\usepackage[pagebackref,breaklinks,colorlinks,citecolor=cvprblue]{hyperref}
\usepackage{graphicx}
\usepackage{booktabs}
\usepackage{multirow}
\usepackage{adjustbox}
\usepackage{bbm}
\usepackage{bm}
\usepackage{indentfirst}
\usepackage{floatrow}
\newfloatcommand{figurebox}{figure}[\nocapbeside][\dimexpr(\textwidth-\columnsep)/2\relax]
\newfloatcommand{tablebox}{table}[\nocapbeside][\dimexpr(\textwidth-\columnsep)/2\relax]

\def\paperID{*****} 
\def\confName{CVPR}
\def\confYear{2024}

\title{SparseFusion: Efficient Sparse Multi-Modal Fusion Framework for Long-Range 3D Perception} 

\author{
Yiheng Li\\
University of Chinese Academy of Sciences\\
{\tt\small liyiheng23\@gmail.com}
\and
Hongyang Li\\
TuSimple\\
{\tt\small lhy\_ustb@pku.edu.cn}
\and
Zehao Huang\\
TuSimple\\
{\tt\small zehaohuang18@gmail.com}
\and
Hong Chang \\
University of Chinese Academy of Sciences\\
{\tt\small changhong@ict.ac.cn}
\and
Naiyan Wang\\
TuSimple\\
{\tt\small winsty@gmail.com}
}

\begin{document}
\maketitle
\input{sec/0_abstract}
\input{sec/1_introduction}
\input{sec/2_related_work}
\input{sec/3_method}
\input{sec/4_experiment}
\input{sec/5_conclusion}

{
    \small
    \bibliographystyle{ieeenat_fullname}
    \bibliography{main}
}

\input{sec/6_suppl}

\end{document}

%% file: preamble.tex
\usepackage[dvipsnames]{xcolor}


%% file: sec/0_abstract.tex
\begin{abstract}
Multi-modal 3D object detection has exhibited significant progress in recent years. However, most existing methods can hardly scale to long-range scenarios due to their reliance on dense 3D features, which substantially escalate computational demands and memory usage. In this paper, we introduce \emph{SparseFusion}, a novel multi-modal fusion framework fully built upon sparse 3D features to facilitate efficient long-range perception. The core of our method is the \emph{Sparse View Transformer} module, which selectively lifts regions of interest in 2D image space into the unified 3D space. The proposed module introduces sparsity from both semantic and geometric aspects which only fill grids that foreground objects potentially reside in. Comprehensive experiments have verified the efficiency and effectiveness of our framework in long-range 3D perception. Remarkably, on the long-range Argoverse2 dataset, SparseFusion reduces memory footprint and accelerates the inference by about two times compared to dense detectors. It also achieves state-of-the-art performance with mAP of 41.2\% and CDS of 32.1\%. The versatility of SparseFusion is also validated in the temporal object detection task and 3D lane detection task. Codes will be released upon acceptance.
\end{abstract}

%% file: sec/1_introduction.tex
\section{Introduction}
\label{sec:intro}

3D object detection, a crucial component in autonomous driving, aims to precisely identify and classify objects in 3D environments. Recent advancements in this field \cite{huang2021bevdet, li2022bevformer, wang2023mv2d, fan2023fsdv2, liu2023bevfusion, wang2023unibev, yan2023cmt, wang2023unitr} typically utilize inputs like surrounding images, sparse point clouds, or multi-modal data, leading to significant enhancements in object localization and categorization accuracy. However, most of these methods are tailored for short-range perception (e.g., 0-50m in nuScenes \cite{caesar2020nuscenes}, 0-75m in Waymo \cite{sun2020waymo}). This limitation becomes evident in practical applications where autonomous vehicles, especially huge and high-speed ones like trucks, require a broader perception range. Consequently, the exploration of long-range perception has become increasingly important and is drawing considerable attention in the field.

Recently, the exploration of long-range perception \cite{chen2023voxelnext, fan2022fsd, fan2023fsdv2} has primarily concentrated on LiDAR point clouds. This preference stems from LiDAR's inherent sparsity over extended ranges, coupled with its capability to accurately localize faraway objects. To circumvent the computational and memory burdens associated with creating dense Bird's Eye View (BEV) features in long-range scenarios, these LiDAR-based 3D detectors \cite{chen2023voxelnext, fan2022fsd, fan2023fsdv2, fan2022sst, pei2023clusterformer} typically first transform point clouds into sparse voxel features \cite{zhou2020MVF, zhou2018voxelnet}, and then adopt sparse convolution \cite{yan2018second, graham20183subm} or Transformer \cite{fan2022sst, liu2023flatformer, wang2023dsvt} to extract features. However, these methods may suffer from performance degradation, especially for small objects, due to the lack of semantic information and diminished point density at extended distances.

\begin{figure}
\centering
\begin{subfigure}{0.45\linewidth} 
    \centering
    \includegraphics[width=\textwidth]{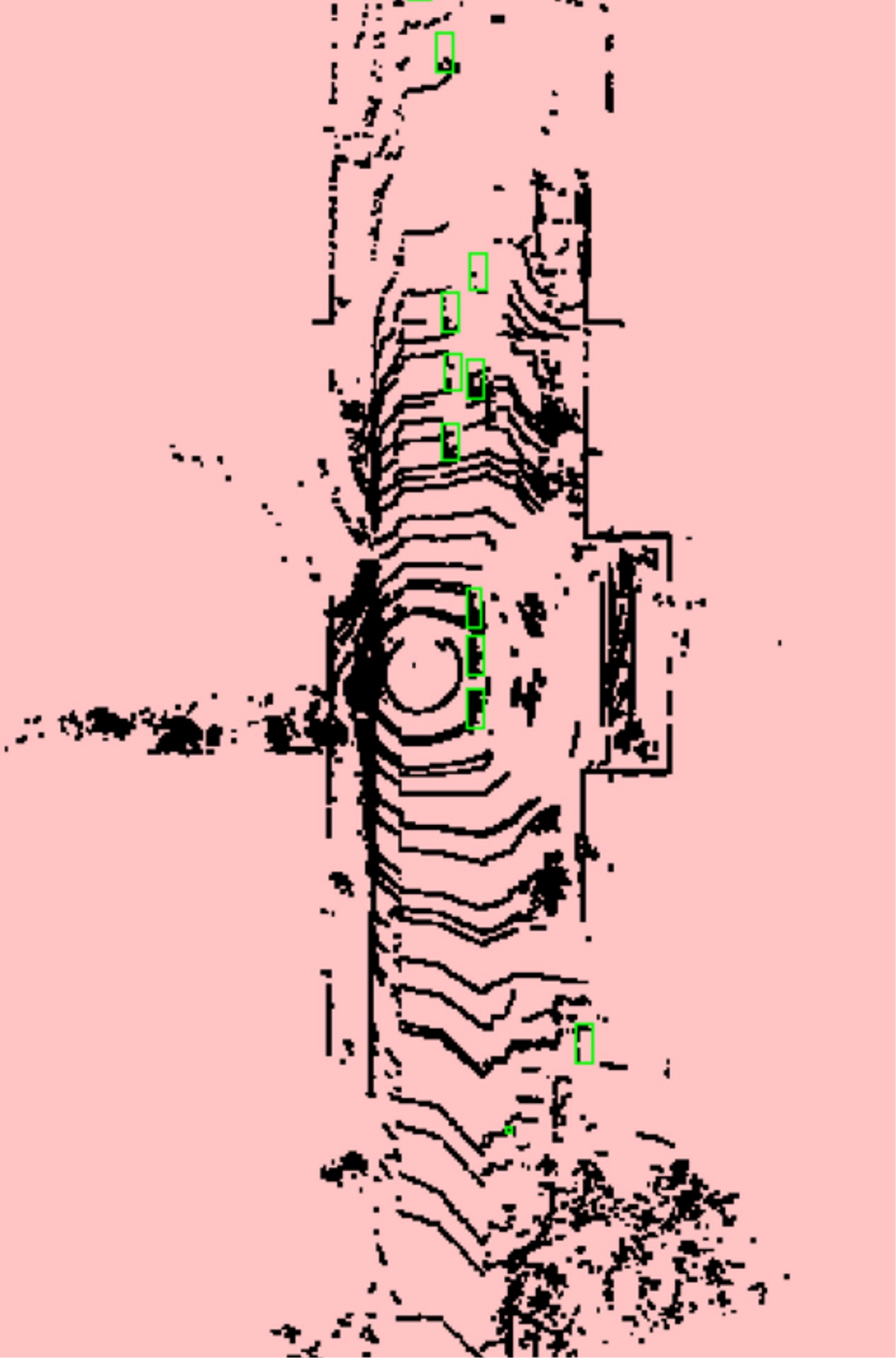}
    \caption{Original LSS}
    \label{fig:heatmap_dense}
\end{subfigure}
\begin{subfigure}{0.45\linewidth} 
    \centering
    \includegraphics[width=\textwidth]{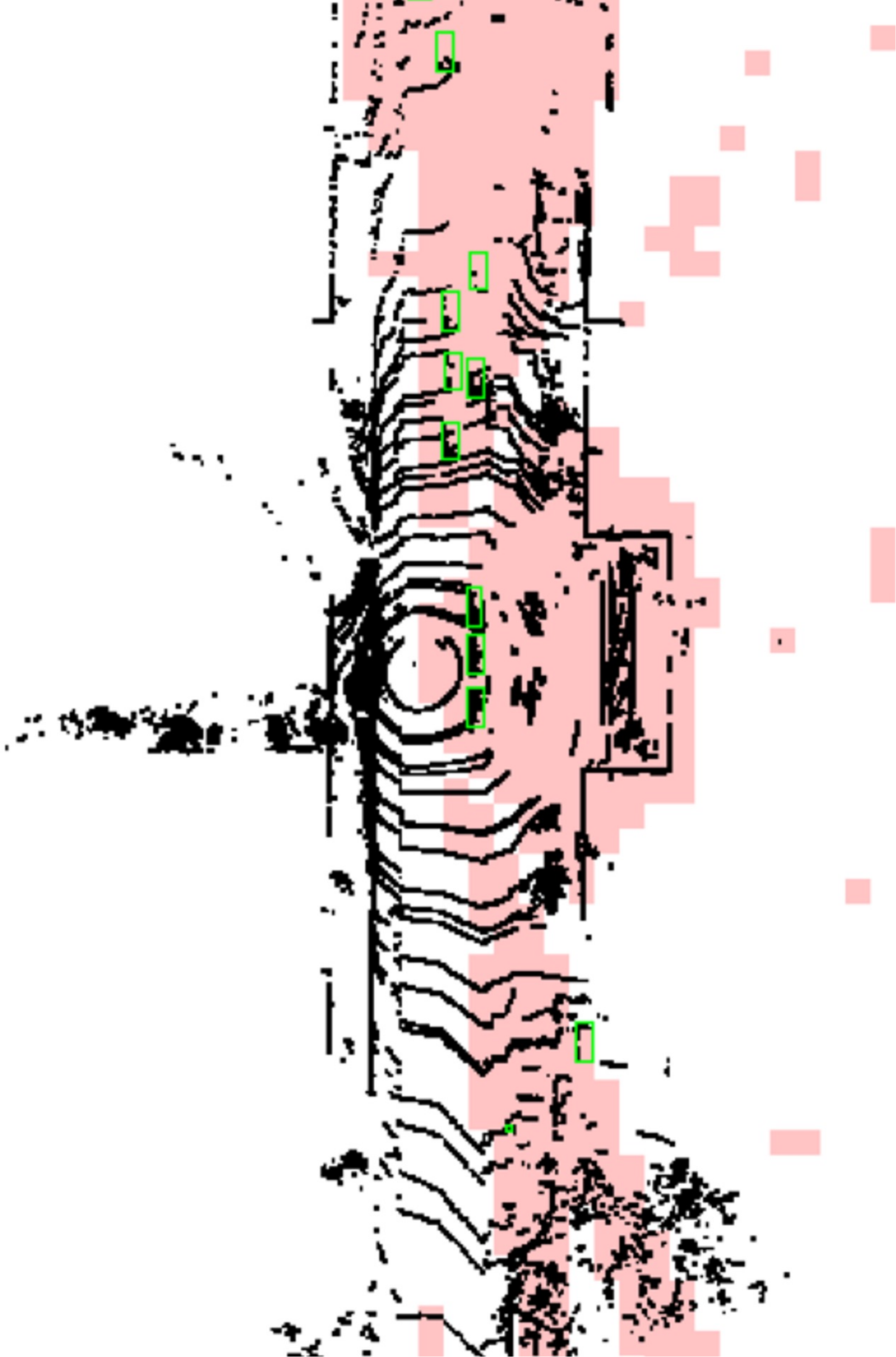}
    \caption{Ours}
    \label{fig:heatmap_sparse}
\end{subfigure}
\caption{Visualization of sparse features generated from Lift-Splat-Shot (LSS) \cite{philion2020lift} and our Sparse View Transformer. The pink area denotes non-empty voxels predicted from images. We filter most of the background in the scene, which consumes less memory and computation resources in long-range perception.}
\label{fig:heatmap}
\end{figure}
Compared with sparse point cloud data, vision data can provide rich and dense semantic information. In recent years, there has been increasing attention to camera-based 3D detections, with a predominant focus on work based on multi-view cameras. These multi-view 3D detectors can be categorized into two classes based on whether they generate BEV feature representations: BEV-based and BEV-free methods. BEV-based methods \cite{huang2021bevdet, li2023bevdepth, li2022bevformer, xie2022m2bev, li2023fastbev, yang2023bevformerv2} explicitly generate dense BEV features and perform 3D detection on this feature. While camera-based BEV methods provide a unified representation space for the fusion of multi-modal features, they encounter analogous challenges to dense BEV-based LiDAR methods, particularly in terms of escalating memory and computation demands with increasing perception ranges. 
Conversely, BEV-free methods pioneered by DETR3D \cite{detr3d}, adopt an alternative approach. Utilizing sparse queries for aggregating image features \cite{liu2022petr, lin2022sparse4d, wang2023streampetr, jiang2023far3d, wang2023mv2d}, they circumvent the need for dense BEV features, thereby allowing for an easy expansion of the perception range. Nevertheless, pure camera-based methods still exhibit a discernible gap in terms of localization accuracy compared to LiDAR-based detectors.

To capitalize on the strengths of different sensors, numerous methods employing multi-modal sensors for 3D detection have emerged \cite{liu2023bevfusion, liang2022bevfusion,yan2023cmt,ge2023metabev, bai2022transfusion, wang2023unibev, wang2023unitr}. These approaches aim to integrate the rich semantic information provided by cameras with the precise localization capabilities of LiDAR. By fusing these diverse data sources, multi-modal 3D detectors can achieve more accurate and reliable object detection performance. However, the prevalent multi-modal fusion methods such as BEVFusion \cite{liu2023bevfusion,liang2022bevfusion} largely depend on dense BEV feature representations, which poses significant challenges when extending these methods to long-range perception. Exploring how to make these BEV-based multi-modal fusion methods work effectively in long-range scenarios is a topic worth investigating.

In this paper, we propose an efficient framework, named SparseFusion, that extends BEV-based methods to long-range perception. Drawing inspiration from the inherent sparsity of point clouds, we focus our attention on specific elements within the 3D environment, such as objects or lanes. By dedicating computational efforts to these critical elements, our framework aims to minimize computational overhead while enabling effective perception over extended ranges. To this end, we propose a Sparse View Transformer module. It selectively lift the 2D information of interest into 3D space by semantic and geometric priors predicted from 2D perception tasks, which results in sparse 3D features. Specifically, we predict the bounding boxes or masks of foreground objects and depth distribution, and only fill those BEV grids the objects may exist. 
These features are then fused with sparse features from the point cloud, leading to the sparse multi-modal 3D features, as shown in Fig.~\ref{fig:heatmap}. Finally, we utilize a deliberately designed sparse feature encoder and heads to generate detection results. Additionally, we also extend our method to temporal detection and 3D lane detection to validate the versatility of our framework.

With the help of Sparse View Transformer, our SparseFusion achieves new state-of-the-art performance on the long-range detection dataset Argoverse2 \cite{Argoverse2}, and it also demonstrates competitive results on the nuScenes dataset. Our main contributions are as follows:
\begin{itemize}
    \item[$\bullet$] We propose a novel multi-modal sparse fusion 3D detection framework, which extends the capabilities of BEV-based methods for long-range perception. In particular, it sparsifies 3D features with the help of semantic and geometric priors from 2D perception tasks.
    \item[$\bullet$] On the long-range perception dataset Argoverse2, our model achieves state-of-the-art performance while remarkably reducing the computational and memory footprint. We also demonstrate the generality of the proposed method on the temporal object detection task and 3D lane detection task.
\end{itemize}

%% file: sec/2_related_work.tex
\section{Related Work}
\label{sec:related_work}
\subsection{LiDAR-based 3D Object Detection} 
LiDAR-based 3D object detection methods \cite{zhou2018voxelnet, lang2019pointpillars, yin2021centerpoint} typically involve dividing the irregular point clouds into regular voxels or pillars \cite{zhou2020MVF, zhou2018voxelnet}, followed by feature extraction using the PointNet \cite{qi2017pointnet} approach. These features, once remapped to a dense feature space, are then processed using dense convolutions for subsequent 3D perception tasks. 
However, these detectors, which rely on dense features, are constrained by the size of the feature space due to finite computing resources. This limitation becomes a significant challenge when attempting to extend their effectiveness to long-range perception, as it necessitates a substantial increase in the size of the dense feature space.

In response to this challenge, SECOND \cite{yan2018second} introduced sparse convolution for feature extraction, effectively reducing both memory and time overhead. Subsequent detectors based on sparse features \cite{yin2021centerpoint, lang2019pointpillars, pei2023clusterformer} further validated the advantages of sparse convolution in 3D detection tasks involving point clouds. Compared with sparse convolution-based methods, point cloud transformer \cite{fan2022sst, liu2023flatformer, wang2023dsvt} enables the capture of long-distance dependencies within data by grouping irregular and sparse point cloud into regular windows and performing self-attention within the windows. 

To further reduce the computational overhead, recent work focus on removing the dense detection head. FSD~\cite{fan2022fsd} first propose a fully sparse structure for LiDAR-based 3D detection by segmentation-clustering-refinement pipeline. VoxelNext~\cite{chen2023voxelnext} improves it by proposing a fully end-to-end architecture through a novel label assignment strategy. Very recently, FSDv2~\cite{fan2023fsdv2} proposes a simple yet effective approach by introducing the concept of ``virtual voxels". All these methods greatly facilitate the long range perception of LiDAR-based methods.

\subsection{Multi-View 3D Object Detection}
Recent advancements in multi-view 3D object detection are focused on directly detecting objects within 3D spaces. These multi-view 3D detection methods can be roughly categorized into two categories: BEV-based methods and BEV-free methods. BEV-based methods \cite{huang2021bevdet, li2023bevdepth, li2022bevformer, li2022bevstereo, xie2022m2bev, zhang2023sabev, liu2023sparsebev} explicitly create dense BEV features through techniques such as LSS \cite{huang2021bevdet, li2023bevdepth, xie2022m2bev, zhang2023sabev} or deformable attention \cite{zhu2020deformable,li2022bevformer, yang2023bevformerv2}, and subsequently perform detection on these features. Nevertheless, as the perception distance increases, the necessity to generate additional BEV grids leads to a quadratic growth in both memory and computational overhead. This challenge makes it difficult to extend BEV-based models for long-range perception.

BEV-free methods avoid the explicit generation of BEV features. Instead, they introduce a set of sparse object queries for feature aggregation and object prediction \cite{detr3d, lin2022sparse4d, liu2022petr, liu2022petrv2, wang2023mv2d, jiang2023far3d}. As a pioneer in this field, DETR3D \cite{detr3d} generates 3D reference points from object queries, leveraging intrinsic and extrinsic parameters to project it onto images, thus facilitating the acquisition and aggregation of multi-view features. PETR \cite{liu2022petr, liu2022petrv2} introduces a novel 3D position-aware feature representation, enabling object queries to directly interact with dense image features through cross-attention. MV2D \cite{wang2023mv2d} capitalizes on the high recall characteristics of 2D detectors to generate dynamic object queries, thereby enhancing performance with fewer object queries. These methods do not generate dense BEV features, which reduces the computational burden for long-range perception. 

\subsection{Multi-modal based 3D Detection}
Multi-modal inputs offer complementary information about the surrounding environment, making multi-modal-based methods capable of achieving superior performance compared to LiDAR-only or vision-only methods. Recent multi-modal 3D detection approaches predominantly rely on BEV feature representations for detection \cite{liu2023bevfusion, bai2022transfusion, chen2023futr3d, hu2023fusionformer, ge2023metabev, hu2023ealss, wang2023unibev}. BEVFusion \cite{liu2023bevfusion} employs the LSS \cite{philion2020lift} operation to project image features into the BEV space, then integrates them with LiDAR features through a straightforward concatenation. TransFusion \cite{bai2022transfusion} and FUTR3D \cite{chen2023futr3d} generate sparse global queries and then refine them through cross-attention with features from LiDAR and images. All of these methods are built on dense feature representations. As the perception distance increases, the computational load and memory requirements of the model experience a significant surge, limiting the practical application of such models in long-range scenarios.


%% file: sec/3_method.tex
\newcommand{\Lcal}{\mathcal{L}}
\newcommand{\Ccal}{\mathcal{C}}

\section{Method}
\label{sec:method}
\begin{figure*}[t]
  \centering
   \includegraphics[width=0.85\textwidth]{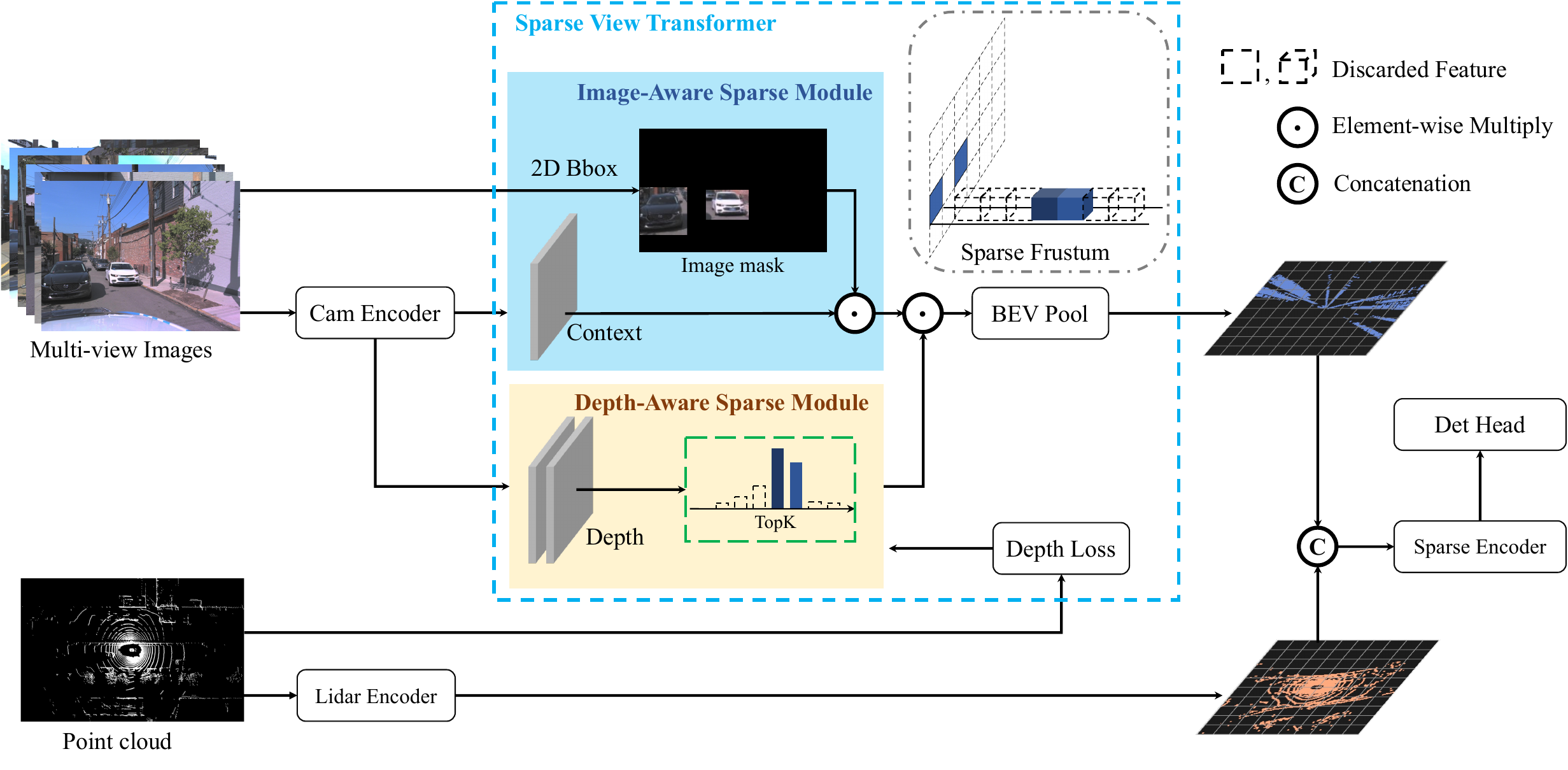}
   \caption{SparseFusion comprises four key modules: the Feature Extractor, Sparse View Transformer, Sparse Encoder, and Detection Head. Initially, the LiDAR point cloud and camera images undergo separate encoders for feature extraction. In the case of camera image features, our Sparse view transformer selectively filters out most of the background information and discards depth values with low confidence, resulting in the generation of sparse 3D features. Subsequently, we fuse features from different modalities and employ a fully sparse encoder for feature extraction. Finally, we apply the detection head to predict the results. }
   \label{fig:framework}
\end{figure*}

The overall architecture of our proposed SparseFusion is illustrated in Fig.~\ref{fig:framework}. We employ two separate networks as feature extractors for camera and LiDAR inputs. The LiDAR branch utilizes a sparse voxel encoder \cite{zhou2018voxelnet} to obtain sparse features $F_{\Lcal} \in \mathbb{R}^{N_{\Lcal}} \times C$, where $N_{\Lcal}$ and $C$ denote the voxel number and the feature dimensions, respectively.
The camera branch takes $N$ surround view images as input, denoted as $I = \{I_1, I_2, ..., I_N\}$, where $I_i \in \mathbb{R}^{H \times W \times 3}$. Then an image backbone (e.g., ResNet \cite{he2016resnet}, Swin \cite{liu2021swin}) followed by a neck (e.g., FPN \cite{lin2017fpn}) is adopted to extract multi-view image features, represented as $F = \{F_1, F_2, ..., F_N\}$, with $F_i \in \mathbb{R}^{H^f \times W^f \times C}$ being the feature extracted from image $I_i$.

After obtaining image features, recent BEV-based methods \cite{huang2021bevdet, li2023bevdepth, li2022bevformer} transform 2D features into dense BEV features using techniques like LSS \cite{philion2020lift} or deformable attention \cite{zhu2020deformable}. However, these methods lead to a substantial increase in computational load and memory requirements, particularly as the perception range expands. To effectively tackle this challenge, we introduce the Sparse View Transformer module, which selectively lifts only the foreground information of interest into 3D space, thereby generating sparse features, represented as $F_{\Ccal} \in \mathbb{R}^{N_{\Ccal} \times C}$\footnote{We omit the difference in feature dimensions between LiDAR features and image features for simplicity.}, where $N_{\Ccal}$ is the number of foreground voxels. 

Once we have derived sparse features from images, we can integrate them with the sparse LiDAR features. This integration is followed by sparse encoders and sparse heads to obtain the final perception results.

\subsection{Sparse View Transformer} \label{sparse-view-transformer}

\begin{figure*}[t]
    \centering
    \begin{subfigure}{0.48\linewidth}  
        \centering
        \includegraphics[width=0.95\linewidth]{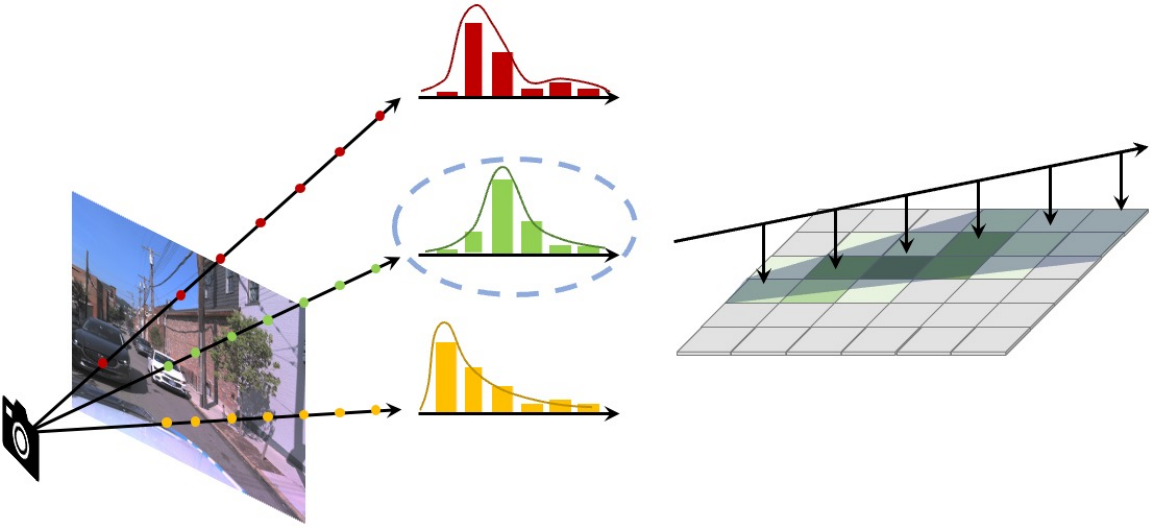}
        \caption{Dense View Transformer}
        \label{fig:original_lss}
    \end{subfigure}
    \begin{subfigure}{0.48\linewidth}  
        \centering
        \includegraphics[width=0.95\linewidth]{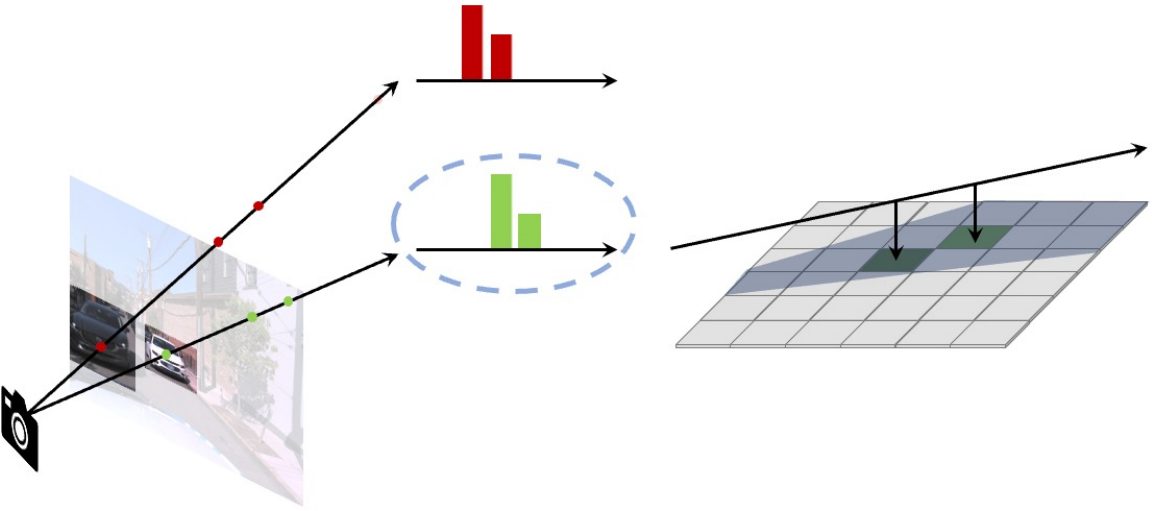}
        \caption{Sparse View Transformer}
        \label{fig:sparse_lss}
    \end{subfigure}
    \caption{Different view transformer module. In (a), depth distribution is predicted for each pixel, and all frustums are projected to BEV. In (b), the background of the image is first filtered out, then the top-K values corresponding to the depth distribution of pixels in the foreground region of the image are retained. This significantly reduces the sparsity of the BEV features generated.}

    \label{fig:lss}
\end{figure*}

Before delving into our proposed Sparse View Transformer, we first provide a brief overview of the LSS \cite{philion2020lift} method, which our approach builds upon. The core idea of LSS is to explicitly predict depth distribution for every image pixel and then lift 2D pixels into 3D space by the predicted depth. For each image pixel, LSS predicts a context vector $\bm{v} \in \mathbb{R}^{C}$ and a depth distribution $\bm{\alpha} \in \mathbb{R}^{|D|}$, where $D$ is a set of discrete depths, as illustrated in Fig.~\ref{fig:original_lss}. Subsequently, the context vector $\bm{v}$ is weighted by $\bm{\alpha}$ to lift each 2D pixel to 3D space, resulting in a feature distribution tensor of the image, denoted as $P_i \in \mathbb{R}^{H^f \times W^f \times C \times |D|}$. Then each $P_i$ is transformed to a unified coordinate system by camera extrinsic, followed by a pillar pooling operation \cite{huang2022bevpoolv2} for multi-view feature fusion. Finally, a dense BEV feature representation is obtained, which is then utilized by feature encoders and specific task heads for further processing.  However, this approach encounters challenges of slow computation and high memory footprint, particularly in long-range perception scenarios which need increasing the number of depth bins $|D|$ and the size of feature map $H^f \times W^f$ for better performance.

To address these challenges, we draw inspiration from the inherent sparsity of point clouds. Our solution revolves around introducing sparsity into the feature tensor, thereby reducing the complexity of the LSS operation and enabling the use of sparse convolution to accelerate the computation of the subsequent BEV feature encoder and head. In the following, we elaborate on how we introduce sparsity into both the spatial and depth dimensions of the feature tensor, corresponding to the Image-Aware Sparse Module and Depth-Aware Sparse Module, respectively.

\noindent\textbf{Image-Aware Sparse Module.} 
To mitigate the computational burden associated with the size of feature map $H^f \times W^f$, our Image-Aware Sparse Module focuses on selectively retaining only the foreground objects, as depicted in Fig.~\ref{fig:sparse_lss}. This is achieved by using 2D detectors \cite{renNIPS15fasterrcnn, redmon2016yolo, tian2019fcos} to identify and generate bounding boxes for each image, represented as $B_\nu = \{x, y, w, h\}$, where $\nu \in \{0, 1, ..., n - 1\}$. These bounding boxes are then employed to generate the foreground mask,
which effectively filters out background pixels, retaining only the foreground objects such as pedestrians, vehicles, and other targets of interest. Then we perform the LSS operation only on the candidate foreground pixels, thereby making the model focus on specific areas of interest in the image. This approach not only reduces the computational complexity of LSS but also furnishes valuable prior information for localizing objects in 3D space.

\noindent\textbf{Depth-Aware Sparse Module.} Another factor impacting the computational complexity of LSS is the number of depth bins $|D|$. With a constant resolution for each depth bin, $|D|$ escalates as the perception distance extends. To tackle the challenge of enumerating a larger number of depth values for long-range perception, we retain only the depth values predicted for each pixel with top-K highest probabilities, while setting the others to zero. This enables us to generate a depth distribution mask for each pixel.

By applying both the image-aware mask and depth-aware mask to $P_i$, we can obtain a sparse tensor for each image. Then we adopt the operation following BEVPoolV2 \cite{huang2022bevpoolv2} to construct a sparse BEV feature which is subsequently utilized by the sparse encoder and head for further processing. Notably, to maximize the sparsity of BEV features derived from images, it is essential that the depth distribution are as concentrated as possible. Taking inspiration from BEVDepth \cite{li2023bevdepth}, we supervise the depth distribution using a ground-truth depth map, which is obtained by projecting the point cloud onto the corresponding image.

\subsection{Sparse Encoder}
After obtaining sparse features from both the camera and LiDAR, a fusion module followed by a sparse encoder is employed for feature fusion and refinement.
Various methods are available for fusing sparse features from different modalities, such as addition, concatenation, and cross-attention \cite{vaswani2017attention}. Here, we adopt concatenation for the fusion of multi-modal features, prioritizing simplicity in our approach. Following BEVFusion \cite{liu2023bevfusion}, we employ a convolution-based BEV encoder to further extract features from the fused sparse tensor. This helps address the issue of spatial misalignment between camera features and LiDAR features, primarily caused by inaccurate depth estimation from images. Notably, we replace the standard convolution used in the original BEV encoder with sparse convolution \cite{yan2018second}.  This modification avoids generating dense feature maps, aligning well with the demands of long-range perception.

\begin{table}
    \centering
    \begin{tabular}{lcc}
        \toprule
        Head&mAP$\uparrow$&CDS$\uparrow$\\
        \midrule
        CenterPoint \cite{yin2021centerpoint}& 0.357 & 0.282\\
        TransFusion \cite{bai2022transfusion}& 0.379 & 0.295\\
        \bottomrule
    \end{tabular}  
    \caption{SparseFusion performance comparison with different heads on AV2.}
    \label{tab:argo-head}
\end{table}

\subsection{Sparse Object Detection Head}
For 3D object detection, our framework accommodates various detection heads, including both convolution-based heads, such as CenterHead from CenterPoint \cite{yin2021centerpoint}, and query-based heads, like the TransFusion head \cite{bai2022transfusion}. The performances of our framework utilizing both CenterHead and TransFusion head on Argoverse2 (AV2) are detailed in Tab.~\ref{tab:argo-head}. 
In our experiments, we primarily employ the TransFusion head, favoring it for its compatibility with our sparse framework and enhanced performance, unless stated otherwise. All the standard convolution operations are replaced with sparse convolution. Furthermore, we adopt deformable attention \cite{zhu2020deformable} instead of global attention. Not only does deformable attention help reduce memory usage and computational overhead, but with an increased perception distance, more background information becomes present in the features, potentially making global attention more susceptible to interference in feature aggregation by the background information.

%% file: sec/4_experiment.tex
\section{Experiment}
\label{experiment}
\subsection{Datasets and Metrics}

We primarily conduct experiments and ablation studies on the Argoverse2 dataset \cite{Argoverse2}, which is particularly suited for long-range perception. Additionally, we also evaluate on the widely recognized nuScenes dataset \cite{caesar2020nuscenes} to facilitate comparisons with various state-of-the-art (SOTA) methods. 

\noindent\textbf{Argoverse2} is a large-scale and long-range dataset, with a perception distance of 200m (cover an area of 400m $\times$ 400m). It contains 1000 scenes, 700 for training, 150 for validation, and 150 for testing. Each scene is recorded with seven high-resolution cameras at 20Hz and one LiDAR at 10Hz. We evaluate SparseFusion with 26 classes within the 200m range. In addition to mean average precision (mAP), the AV2 dataset proposes a comprehensive metric named composite detection score (CDS), taking both AP and localization errors into account.

\begin{table*}
  \centering
      \begin{tabular}{l|cc|cc|ccc}
        \toprule
        Methods & Modality & Voxel size &mAP $\uparrow$ & CDS $\uparrow$ & mATE $\downarrow$ & mASE $\downarrow$ & mAOE$\downarrow$ \\
        \midrule
        Far3D \cite{jiang2023far3d} & C  & - & 0.316 & 0.239 & 0.732 & 0.303 & \textbf{0.459}\\
        \midrule
        CenterPoint \cite{yin2021centerpoint} & L & (0.2, 0.2) & 0.220 & 0.176 & - & - & -\\
        FSD \cite{fan2022fsd} & L & (0.2, 0.2) &0.282 & 0.227 & 0.414 & 0.306 & 0.645\\
        VoxelNext \cite{chen2023voxelnext} & L & (0.1, 0.1) & 0.307 & - & - & - & -\\
        FSDv2 \cite{fan2023fsdv2} & L & (0.2, 0.2) & 0.376 & 0.302 & \textbf{0.377} & \textbf{0.282} & 0.600\\
        \midrule
        FSF \cite{li2023fsf} & CL & (0.2, 0.2) & 0.332 & 0.255 & 0.442 & 0.328 & 0.668\\
        CMT \cite{yan2023cmt} & CL & - & 0.361 & 0.278 & 0.585 & 0.340 & 0.614 \\
        BEVFusion$\dagger$ \cite{liu2023bevfusion} & CL & (0.075, 0.075) &0.388 & 0.301 & 0.450 & 0.336 & 0.643 \\
        \rowcolor{gray!30} SparseFusion & CL & (0.075, 0.075) & \textbf{0.398} & \textbf{0.310} & 0.449 & 0.308 & 0.677\\
        \bottomrule
      \end{tabular}
      \caption{3D object detection results on AV2 val set. $\dagger$ means implemented by ourselves. Bold indicates the best performance. }
      \label{tab:argo2-all}
\end{table*}

\begin{table*}
    \centering
    \begin{tabular}{l|cc|cc|cc}
        \toprule
        Methods & Modality & Backbone & mAP(\textit{test}) & NDS(\textit{test}) & mAP(\textit{val}) & NDS(\textit{val})\\
        \midrule
        PETR \cite{liu2022petr}          & C & Res-101 & 0.391 & 0.455 & 0.370 & 0.442 \\
        BEVDet \cite{huang2021bevdet}    & C & Swin-B & 0.422 & 0.482 & 0.393 & 0.472 \\
        BEVDepth \cite{li2023bevdepth}   & C & VovNet & 0.503 & 0.600 & - & - \\
        M2BEV \cite{xie2022m2bev}        & C & ResNeXt-101 & 0.429 & 0.474 & 0.417 & 0.470 \\
        BEVFormer \cite{li2022bevformer} & C & Res-101 & 0.445 & 0.535 & 0.416 & 0.517 \\
        MV2D \cite{wang2023mv2d}         & C & Res-101 & 0.483 & 0.573 & 0.471 & 0.561 \\
        \midrule
        CenterPoint \cite{yin2021centerpoint} & L & VoxelNet & 0.603 & 0.673 & 0.596 & 0.668 \\ 
        VoxelNext \cite{chen2023voxelnext} & L & - & 0.645 & 0.700 & 0.600 & 0.671 \\
        FSDv2 \cite{fan2023fsdv2} & L & - & 0.662 & 0.717 & 0.647 & 0.704 \\
        \midrule
        Transfusion \cite{bai2022transfusion} & CL & DLA34\&VoxelNet & 0.689 & 0.716 & 0.675 & 0.713\\
        BEVFusion \cite{liu2023bevfusion}     & CL & Swin-T\&VoxelNet & \textbf{0.702} & \textbf{0.729} & 0.685 & \textbf{0.714}\\
        \rowcolor{gray!30} SparseFusion       & CL & Swin-T\&VoxelNet & 0.701 & 0.727 & \textbf{0.687} & 0.706 \\
        \bottomrule
    \end{tabular}
    \caption{3D object detection results on nuScenes(val and test set). Bold indicates the best performance. }
    \label{tab:nusc}
\end{table*}

\begin{table*}
  \centering
  \begin{tabular}{c|l|ccc|ccc|ccc}
    \toprule
     Method & DA & IAS & DAS & mAP & CDS & $S_{fuse}$ & Latency (ms) & Latency* (ms) & Mem (MB) \\
    \midrule
    BEVFusion$\dagger$ & & & & 0.370 & 0.283 & 0 & 361.23 & 270.43 & 6530\\
    BEVFusion$\dagger$ & \checkmark & & & 0.388 & 0.301 & 0 & 327.36 & 249.45 & 1951 \\ 
    \midrule
    SparseFusion & \checkmark & \checkmark & & 0.390 & 0.304 & 49.9\% & 312.50 & 244.51 & 1972 \\
    SparseFusion & \checkmark & & \checkmark & 0.392 & 0.307 & 91.9\% & 151.52 & 84.72 & 1101 \\
    SparseFusion & \checkmark & \checkmark & \checkmark & \textbf{0.398} & \textbf{0.310} & \textbf{93.8\%} & \textbf{140.85} & \textbf{74.17} & \textbf{999} \\
    \bottomrule
  \end{tabular}
  \caption{Ablation of our sparse components on Argoverse2 val set. We utilize BEVFusion as our baseline and progressively integrate additional components: Deformable Attention (DA), Image-Aware Sparse Module (IAS), and Depth-Aware Sparse Module (DAS). * means the latency excluding image and LiDAR backbones. $\dagger$ means we use SyncBatchNorm during training. Bold indicates the best performance. }
  \label{tab:ablation}
\end{table*}

\begin{table*}
  \centering
    \begin{tabular}{l|c|ccc|c|c}
    \toprule
    Method  &  mAP  & V-Transformer & Feat-Encoder & Head & FPS & Mem\\
    \midrule
    BEVFusion$\dagger$ & 0.408 & 525.88        & 47.41        & 23.36 & 1.3 & 6180\\
    SparseFusion & \textbf{0.412} & \textbf{90.21} \textcolor{ForestGreen}{$\downarrow$82.80\%} & \textbf{12.12} \textcolor{ForestGreen}{$\downarrow$74.43\%} & \textbf{15.20} \textcolor{ForestGreen}{$\downarrow$34.93\%} & \textbf{3.8} & \textbf{1817}\\
    \bottomrule
    \end{tabular}
  \caption{Comparison of the latency (ms), FPS (img/s), and memory cost (MB) between our method and baseline with an image resolution of 720 $\times$ 1440 on AV2 val set. We do not list the latency of the image and LiDAR backbone since our model shares the same backbones as baseline methods. We calculated the ratio of latency reduction ratio of SparseFusion compared to the baseline. Bold indicates the best performance. }
  \label{tab:argo-latency}
\end{table*}

\begin{table}
    \centering
    \begin{tabular}{c|c|cccc}
        \toprule
        K & DL & mAP & CDS & $S_{cam}$ & $S_{fuse}$ \\
        \midrule
        \multirow{2}{*}{10} &  & 0.393 & 0.307 & 90.7\% & 89.0\% \\
                            & \checkmark & \textbf{0.394} & \textbf{0.307} & \textbf{95.5\%} & \textbf{93.8\%} \\
        \midrule
        \multirow{2}{*}{20} &  & \textbf{0.396} & 0.309 & 82.7\% & 81.1\% \\
                            & \checkmark & 0.395 & \textbf{0.309} & \textbf{93.9\%} & \textbf{92.4\%} \\
        \midrule
        \multirow{2}{*}{30} &  & 0.394 & 0.307 & 76.1\% & 74.6\% \\
                            & \checkmark & \textbf{0.395} & \textbf{0.308} & \textbf{91.9\%} & \textbf{90.4\%} \\
        \bottomrule
    \end{tabular}
    \caption{The performance of different K for depth values on AV2 val set. DL means depth loss. Bold indicates the best performance. }
    \label{tab:argo2-topk}
\end{table}

\begin{table*}
  \centering
  \begin{tabular}{l|c|c|cccccc}
    \toprule
    Methods & Modality & F-Score $\uparrow$ & X error near $\downarrow$ & X error far $\downarrow$ & Z error near $\downarrow$ & Z error far $\downarrow$ \\
    \midrule
    3DlaneNet \cite{garnett20193dlanenet} & C & 0.441 & 0.479 & 0.572 & 0.367 & 0.443\\
    PersFormer \cite{chen2022persformer} & C & 0.505 & 0.485 & 0.553 & 0.364 & 0.431\\
    Anchor3DLane \cite{huang2023anchor3dlane} & C & 0.537 & 0.276 & 0.311 & 0.107 & 0.138\\
    BEVLaneDet \cite{wang2022bevlanedet} & C & 0.584 & 0.309 & 0.659 & 0.244 & 0.631\\
    BEVLaneDet$\dagger$ \cite{wang2022bevlanedet} & C & 0.613 & 0.446 & 0.431 & 0.115 & 0.145\\
    PETRv2 \cite{liu2022petrv2} & C & 0.612 & 0.400 & 0.573 & 0.265 & 0.413\\
    LATR \cite{luo2023latr} & C & 0.619 & \textbf{0.219} & \textbf{0.259} & \textbf{0.075} & \textbf{0.104}\\
    \midrule
    SparsePoint \cite{yao2023sparsepoint} & CL & 0.523 & 0.468 & 0.514 & 0.371 & 0.418\\
    M$^2$-3DLaneNet \cite{2022arXivm2-3dlane} & CL & 0.555 & 0.431 & 0.487 & 0.327 & 0.401\\
    \rowcolor{gray!40} SparseFusion(Ours) & CL & \textbf{0.638} & 0.413 & 0.403 & 0.083 & 0.123\\
    \bottomrule
  \end{tabular}
  \caption{3D lane detection results on OpenLane val set. $\dagger$ means implemented by ourselves. Bold indicates the best performance. }
  \label{tab:openlane}
\end{table*}

\noindent\textbf{nuScenes} is one of the most widely adopted autonomous driving dataset for 3D object detection. It contains 1000 scenes and each scene lasts for 20 seconds. There are 1.4 million annotated 3D bounding boxes from 10 categories. The perception range of nuScenes is 50m (cover an area of 100m $\times$ 100m). We use the mAP and the nuScenes detection score (NDS) as our detection metrics. 

\subsection{Implementation Details}\label{sec:implementationdetails}
To extract the point cloud features, we adopt VoxelNet \cite{zhou2018voxelnet} as the LiDAR backbone. We set voxel size to (0.075m, 0.075m, 0.2m), and BEV grid size to (0.6m, 0.6m) following BEVFusion \cite{liu2023bevfusion}. For nuScenes dataset, Swin-T \cite{liu2021swin} is used as the image backbone and the input resolution of images is 256$\times$704. We retain the top 10 depth values out of 118 depth bins. As for AV2, we employ ResNet50 \cite{he2016resnet} as the image backbone and resize the source images to 384$\times$768. Top 10 depth values are selected from 648 depth bins. During training, SparseFusion uses ground-truth 2D bounding boxes to generate image masks. The 2D bounding boxes generated by the 2D detector are only used during the inference phase. By default, we use Faster-RCNN \cite{renNIPS15fasterrcnn} with a ResNet50 backbone as the 2D detector. We project the 3D bounding boxes onto the images to obtain 2D ground-truth bounding boxes for the training of Faster-RCNN.

\subsection{Main Results}
To validate the performance of our framework in long-range scenarios, we conduct experiments on the AV2 dataset. We compare our model with existing state-of-the-art algorithms \cite{jiang2023far3d, chen2023voxelnext, fan2023fsdv2, li2023fsf, yan2023cmt} on the AV2 val set. As shown in Tab.~\ref{tab:argo2-all}, our model exhibits remarkable superiority. Compared to our baseline BEVFusion \cite{liu2023bevfusion}, our method achieves 1.0\% mAP and 0.9\% CDS improvements with the same image and voxel resolution. Our model also performs well even in close-range scenarios, as shown in Tab.~\ref{tab:nusc}. Our model achieves similar results to the baseline on the nuScenes dataset.

\subsection{Ablation Studies \& Analyses} \label{ablation}
We conduct ablation studies on the AV2 dataset, following the same implementation as our main experiments in Sec. \ref{sec:implementationdetails}, unless specifically stated.

\noindent\textbf{Sparse Modules.} We present a comprehensive analysis of the sparse components of our model. Starting from BEVFusion \cite{liu2023bevfusion}, we report the performance of each module and then calculate the sparsity of the fused BEV feature (noted as $S_{fuse}$), the latency and memory cost in each setting respectively\footnote{We use torch.cuda.max\_memory\_allocated for the calculation of memory overhead. Sparsity is calculated by dividing the number of empty grids in the feature map by the total number of grids.}.
To ensure a fair comparison with BEVFusion, we substitute the global attention in BEVFusion with deformable attention, using this as the baseline for our sparse module. As shown in Tab.~\ref{tab:ablation}, the exclusive integration of the Image-Aware Sparse Module increases sparsity to 50\%, resulting in a similar performance compared to the baseline. However, due to the inherent overhead of sparse convolution \cite{yan2018second}, a sparsity of 50\% does not result in a significant speed improvement. Readers can refer to the supplementary material for an intuitive understanding of the sparsity-latency/memory relationship. When employing only the Depth-Aware Sparse Module, we achieve a 0.4\% mAP and 0.6\% CDS improvement at a 91.9\% sparsity level compared to the baseline, accompanied by a reduction in latency of nearly 170ms. This emphasizes that directly lifting redundant information from 3D features in long-range scenarios can be excessive. Eliminating this redundancy not only reduces memory usage and computational load but also improves the model's performance. With the incorporation of both IAS and DAS, our framework outperforms the baseline, achieving a 57\% reduction in latency and a 49\% decrease in memory usage. To demonstrate the efficiency of our model more clearly, a detailed breakdown of the latency of SparseFusion and BEVFusion with larger input resolution is provided in Tab.~\ref{tab:argo-latency}.

As shown in Tab.~\ref{tab:argo-latency}, our algorithm reduces the inference latency of about 80\% in view-transformer, 75\% in feature-encoder, and 35\% in detection head compared to the baseline model. By leveraging sparse features, our approach effectively reduces memory usage and enhances inference speed. The improved resource footprint and superior performance demonstrate the suitability of SparseFusion for long-range detection tasks.

\noindent\textbf{Top-K Depth Values.} In our proposed Depth-Aware Sparse Module, we control the sparsity of the output feature by adjusting the top-K values. We conducted ablation experiments on the AV2 validation set with varying K values, as detailed in Tab.~\ref{tab:argo2-topk}. Alongside mAP and CDS, we offer insights into the sparsity of the camera output feature (noted as $S_{cam}$) and the fused feature (noted as $S_{fuse}$). Based on experimental results, even when retaining only the top-10 depths, the model exhibits comparable mAP to the baseline despite the image features having a sparsity of 93.8\%. 

\begin{table}
    \centering
    \begin{tabular}{c|ccc}
        \toprule
        Res & mAP$\uparrow$&CDS$\uparrow$&$S_{fuse}$\\
        \midrule
        384 $\times$ 768  &0.398&0.310&93.8\%\\
        576 $\times$ 1152 &0.406&0.316&93.1\%\\ 
        720 $\times$ 1440 &0.412&0.321&92.3\%\\ 
        \bottomrule
    \end{tabular}
    \caption{The performance of different image resolutions on AV2 val set.}
    \label{tab:argo-img-res}
\end{table}

\begin{figure*}
    \centering
    \begin{subfigure}{0.49\textwidth}
        \centering
        \includegraphics[width=\linewidth]{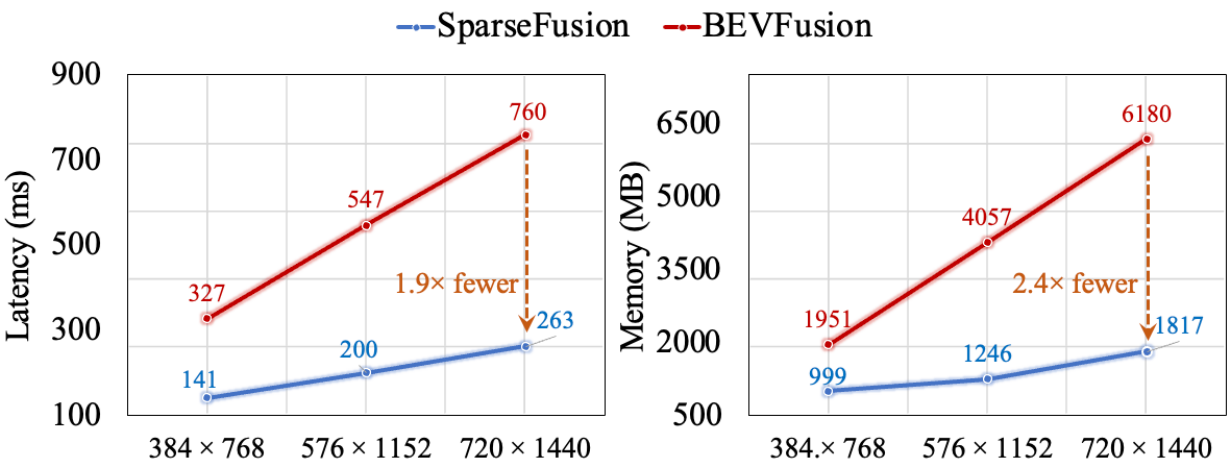}
        \caption{Different image resolutions.}
        \label{fig:img_res_latency_memory}
    \end{subfigure}
    \begin{subfigure}{0.49\textwidth}
        \centering
        \includegraphics[width=\linewidth]{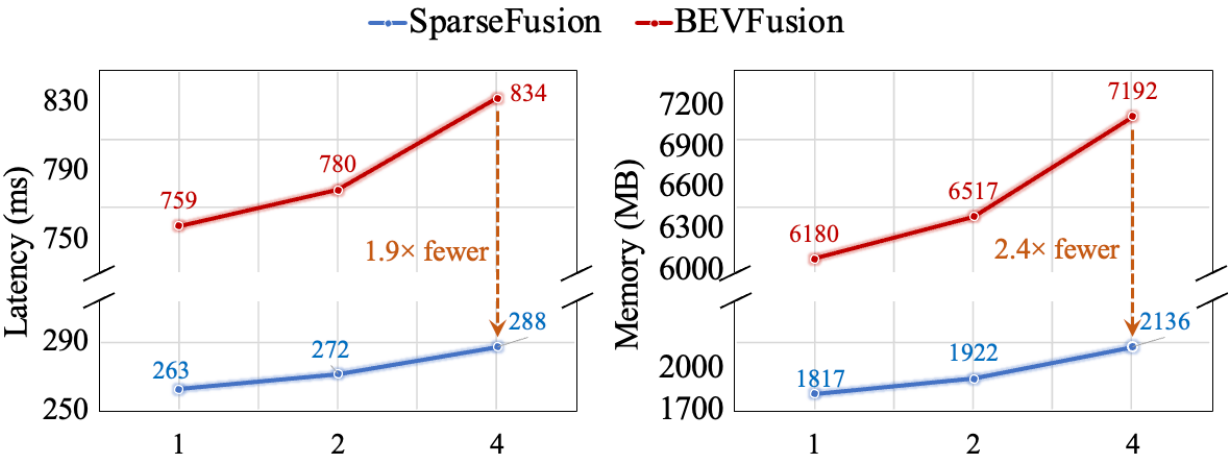}
        \caption{Different history frame counts. }
        \label{fig:temp_latency_memory}
    \end{subfigure}
    \caption{Comparison of the latency and memory cost between our method and BEVFusion. }
\end{figure*}

This suggests that there is a substantial amount of redundant information in the lifted image features, and disregarding this information reduces memory usage and computational overhead. After incorporating depth loss, there is an approximately 5\% to 16\% increase in the sparsity of image features, indicating that the addition of depth loss concentrates the predicted depths, further increasing the model's sparsity.

\noindent\textbf{Input Resolution.} The density of LiDAR points diminishes with increasing distance, which limits the information available for distant objects. In our fusion framework, we address this by increasing the resolution of the image to compensate. Tab.~\ref{tab:argo-img-res} presents the results of our model on the AV2 val set with varying image resolutions. As the image resolution increases, we observe notable improvements in both mAP and CDS, progressing from 39.8\% and 31.0\% to 41.2\% and 32.1\%, respectively. This enhancement in performance is largely due to the higher image resolutions providing clearer visibility of distant objects, thereby offering richer information essential for effective long-range object detection. However, for our baseline method BEVFusion, an increase in image resolution results in a significant rise in latency, as depicted in Fig.~\ref{fig:img_res_latency_memory}. This underscores the necessity of sparsity in our framework: while higher resolution images offer more detailed information, sparse view transformer can efficiently retain necessary information without incurring prohibitive latency.

\noindent\textbf{2D Object Detectors.} We analyze the impact of 2D detectors on the model's performance, as shown in Tab.~\ref{tab:nusc-detector}. When employing Faster-RCNN as the 2D Detector, the model can achieve a 0.6\% mAP improvement. With the use of a relatively weaker 2D Detector (FCOS), the model's performance can still maintain stability, which demonstrates that our model has good robustness to 2D bounding boxes with different qualities.We also provide an oracle experiment that adopts ground-truth 2D bounding boxes as inputs. 

\begin{table}
    \centering
    \begin{tabular}{l|cccc}
            \toprule
            2D Detector & mAP (2D) & mAP (3D) & CDS \\
            \midrule
            None        & -        & 0.392    & 0.307 \\
            Faster-RCNN & 0.339    & 0.398    & 0.310 \\
            FCOS        & 0.290    & 0.396    & 0.309 \\
            GT          & -        & 0.412    & 0.321 \\
        \bottomrule
      \end{tabular}
    \caption{Performance of different 2D detectors on AV2 val set. None means only the Depth-Aware Sparse module (DAS) is retained. GT means using GT 2D bounding box as input to the Image-Aware Sparse module (IAS). }
    \label{tab:nusc-detector}
\end{table}

\begin{table}[]
    \centering
    \begin{tabular}{c|cc|ccc}
        \toprule
        & \multicolumn{2}{c|}{BEVFusion$\dagger$} & \multicolumn{3}{c}{SparseFusion} \\
        \midrule
        \#Frames & mAP$\uparrow$ & CDS$\uparrow$ & mAP$\uparrow$& CDS$\uparrow$ & $S_{fuse}\uparrow$\\
        \midrule
        1 & 0.408& 0.317         & 0.412        & 0.321    & 92.3\% \\
        2 & 0.411& 0.319         & 0.417        & 0.327    & 89.6\% \\
        4 & 0.419& 0.326         & 0.427        & 0.336    & 86.2\% \\
        \bottomrule
        \end{tabular}
    \caption{Temporal 3D detection results on AV2 val set. $\dagger$ means implemented by ourselves. \#Frames denotes the history frame count used for training. }
    \label{tab:temporal}
\end{table}

This oracle experiment achieves higher performance than other detectors, which demonstrates the potential of our SparseFusion framework. We will try to incorporate better 2D priors in future work. 

\subsection{Application to Temporal Object Detection}

Temporal inputs often introduce more redundant information. To assess the efficiency and performance of our model with multi-frame inputs, we extend it to temporal object detection. Following BEVDet4D \cite{huang2022bevdet4d}, we first align the history BEV feature to the current timestamp according to ego-motion, then employ concatenation and sparse convolution approach to merge historical frames. As shown in Tab. \ref{tab:temporal}, as the number of input frames increases, our model consistently outperforms BEVFusion by 0.4, 0.5, and 0.8 mAP respectively, with only 6.1\% decrease in feature sparsity, demonstrating its ability to effectively filter out redundant information. In contrast, temporal detection methods relying on dense BEV representations \cite{huang2022bevdet4d, li2022bevformer, han2023videobev, cai2023bevfusion4d} require the storage of historical frames, and the dense alignment operation becomes a latency bottleneck, particularly in distant perception scenarios. Since our model only fuses foreground regions, it is efficient in terms of both memory and computation, as illustrated in Fig. \ref{fig:temp_latency_memory}. For the 4-frame setting, our model can accelerate inference speed by 1.9 times and reduce memory consumption by 2.4 times.

\subsection{Application to 3D Lane Detection}

To demonstrate the versatility of SparseFusion, we apply it to 3D lane detection. Previous methods \cite{2022arXivm2-3dlane, wang2022bevlanedet, garnett20193dlanenet} typically rely on dense BEV features, which pose challenges in terms of computational overhead for long-range perception. We adapt this method for 3D lane detection by substituting the bounding box-based masks with the masks that come from 2D lane segmentation. As for the head of this task, we adopt the head in BEV-LaneDet \cite{wang2022bevlanedet}.

We conducted 3D lane detection task on OpenLane \cite{chen2022persformer}, a large-scale 3D lane detection dataset built based on the Waymo \cite{sun2020waymo}, and compare our SparseFusion with recent SOTA methods. The results are provided in Tab.~\ref{tab:openlane}. Comparing to the baseline BEV-LaneDet, our method achieved 2.5\% higher F1-score, and both X and Z errors decreased. This improvement underscores the value of precise spatial information from LiDAR points in enhancing localization accuracy. The results also demonstrate the strong compatibility of our model with different tasks. Implementation details are provided in the supplementary material.

%% file: sec/5_conclusion.tex
\section{Conclusion} 

In this paper, we introduce an efficient sparse multi-modal fusion framework, SparseFusion, for long-range 3D perception. Our framework takes advantage of the deliberately designed Sparse View Transformer module, which selectively lifts the region of interest into 3D space and seamlessly integrates it with 3D point cloud features. This approach significantly reduces computational demands and memory usage, extending the capabilities of BEV-based methods for efficient long-range perception. Experimental results highlight the excellent performance of our approach, underscoring its potential for practical applications.

%% file: sec/6_suppl.tex
\clearpage
\setcounter{page}{1}
\maketitlesupplementary

\section*{Implementation details}

\begin{figure*}[htbp]
\centering
\begin{floatrow}[2]
\figurebox[0.3\textwidth]{
    \caption{The BEV sparsity of SparseFusion under different ranges.}
    \label{fig:range_sparsity}}{
        \centering
        \includegraphics[width=0.9\linewidth]{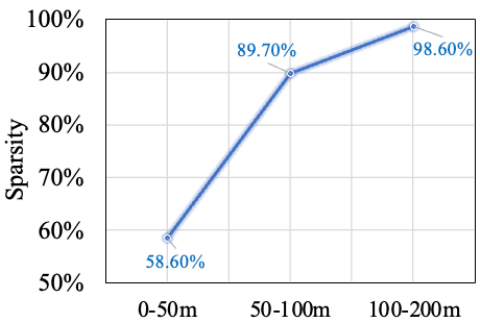}
    }
\figurebox[0.67\textwidth]{
    \caption{The memory cost and latency (exclude the latency of backbones) of SparseFusion under different BEV sparsity. The red lines denote the overhead of BEVFusion \cite{liu2023bevfusion} with deformable attention.}
    \label{fig:sparsity_latency_mem}}{
        \centering
        \includegraphics[width=\linewidth]{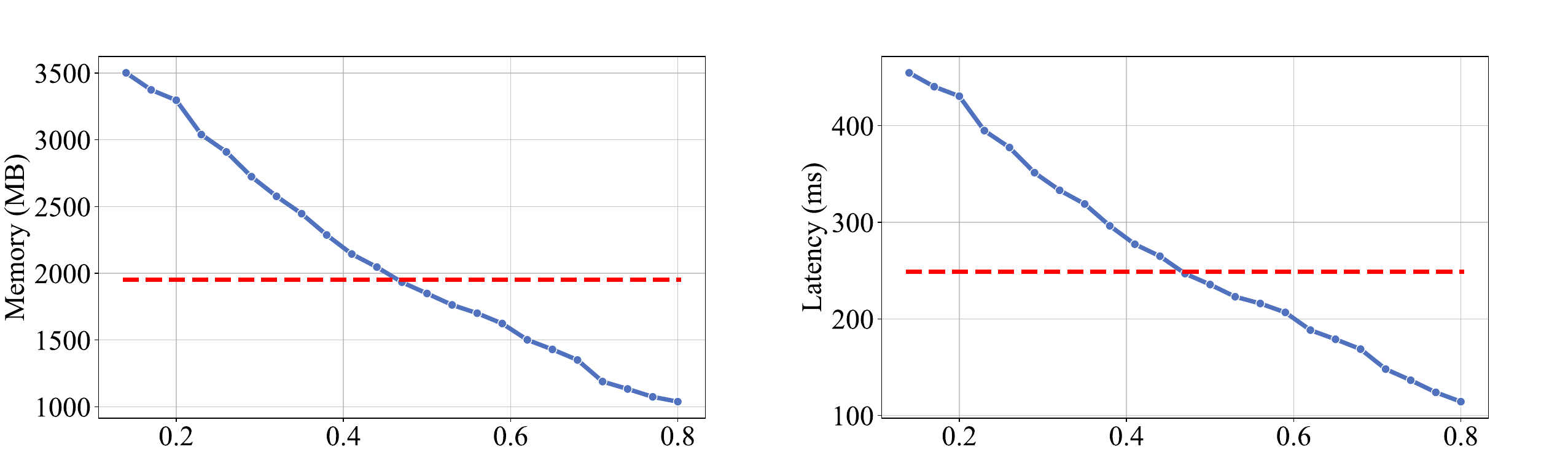}
    }
\end{floatrow}
\end{figure*}

\begin{figure*}[t]
  \centering
   \includegraphics[width=0.19\linewidth]{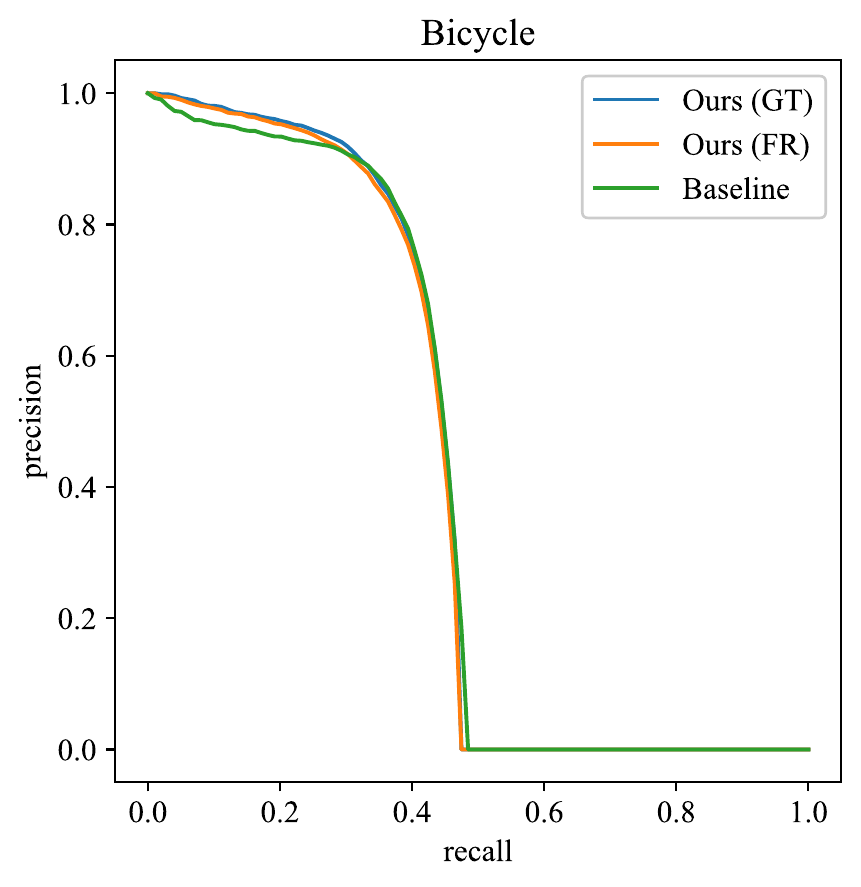}
   \includegraphics[width=0.19\linewidth]{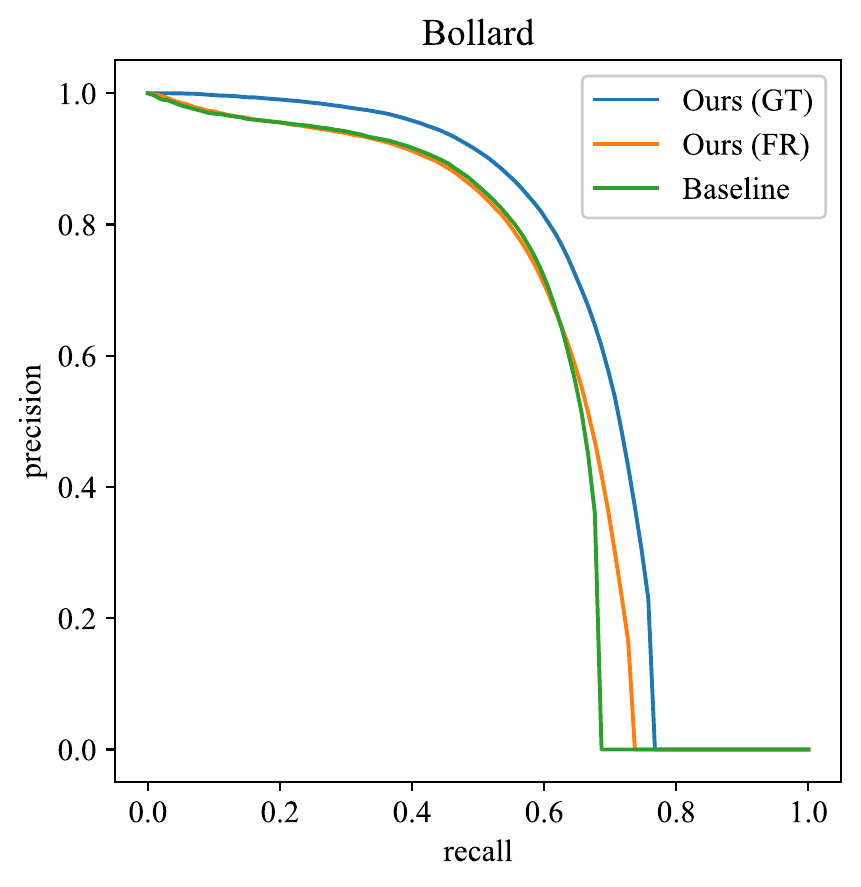}
   \includegraphics[width=0.19\linewidth]{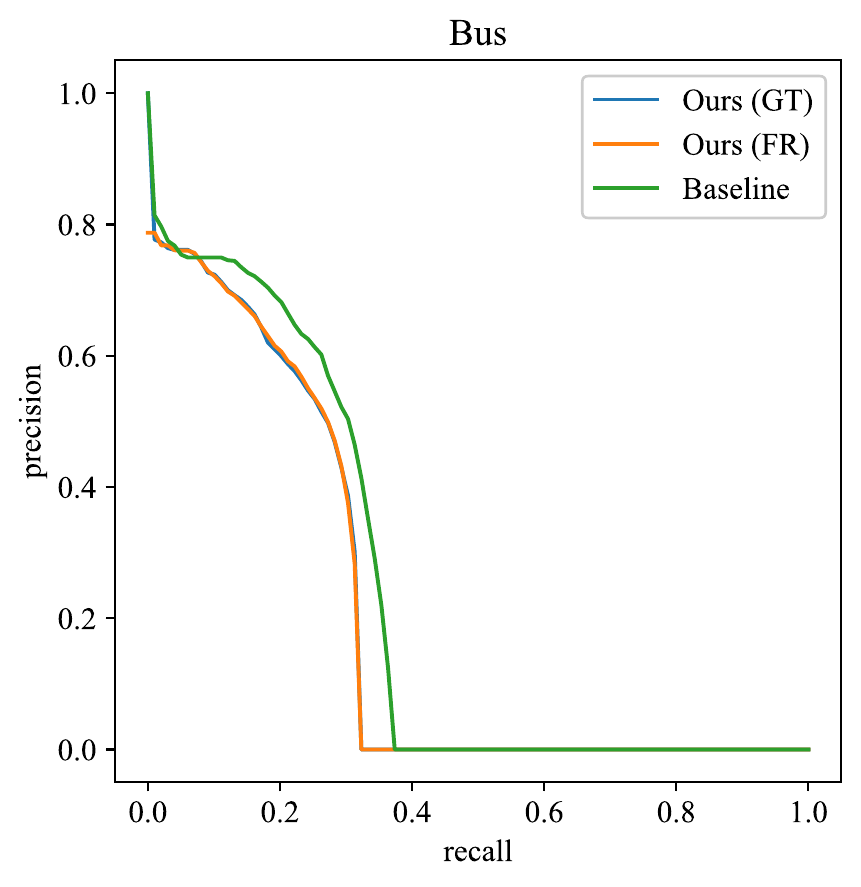}
   \includegraphics[width=0.19\linewidth]{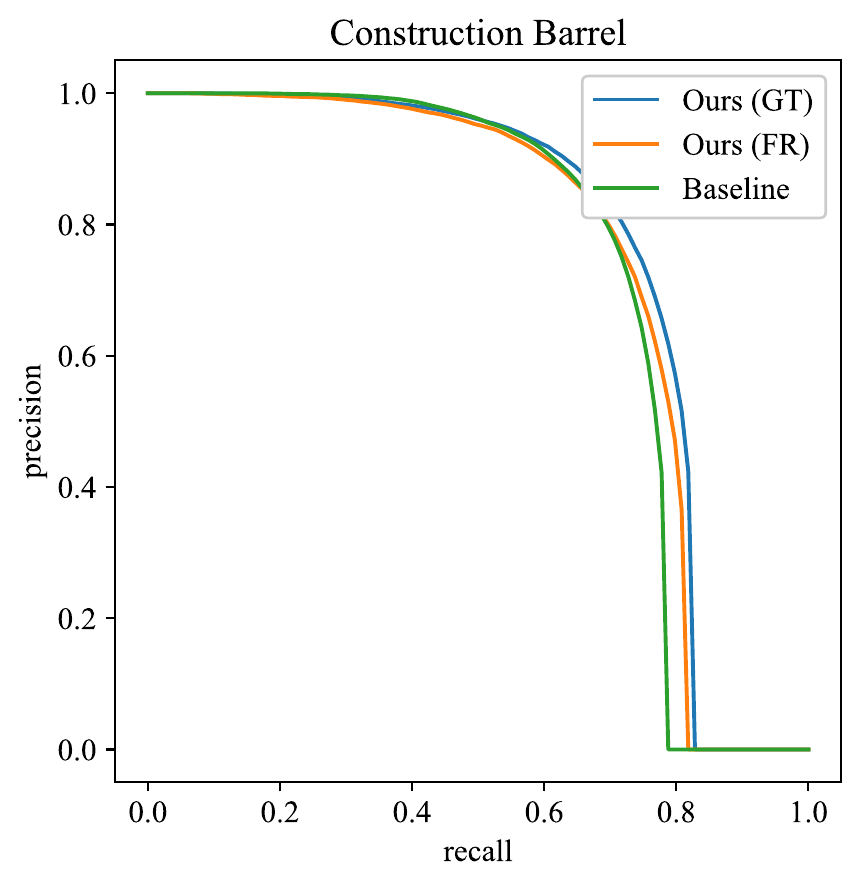}
   \includegraphics[width=0.19\linewidth]{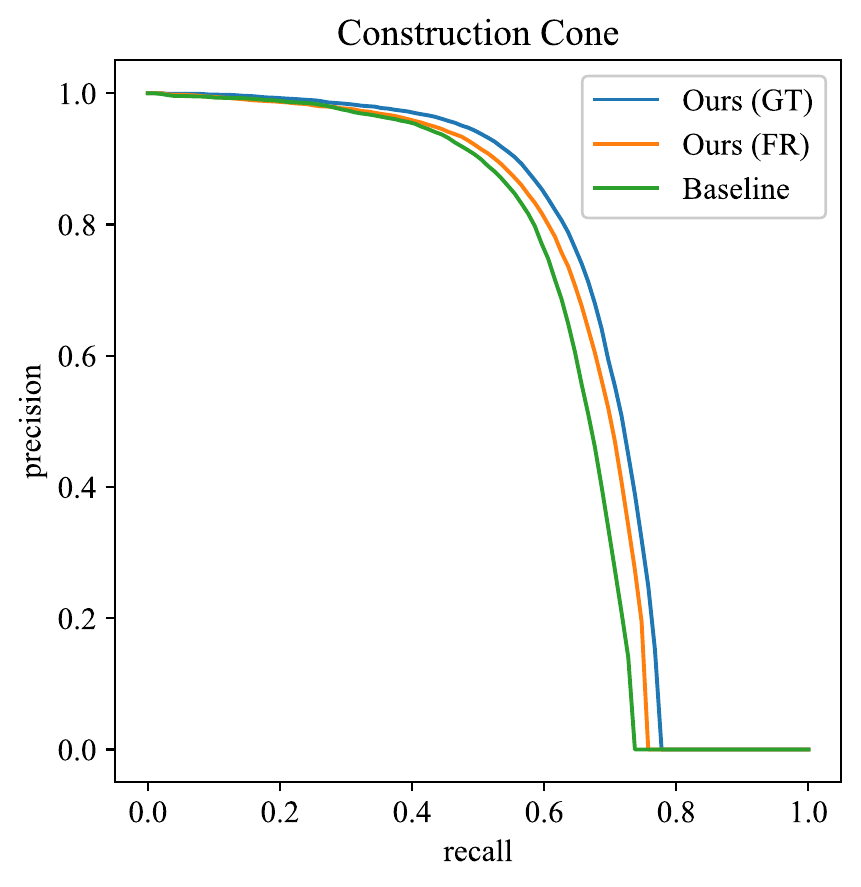}
   
   \includegraphics[width=0.19\linewidth]{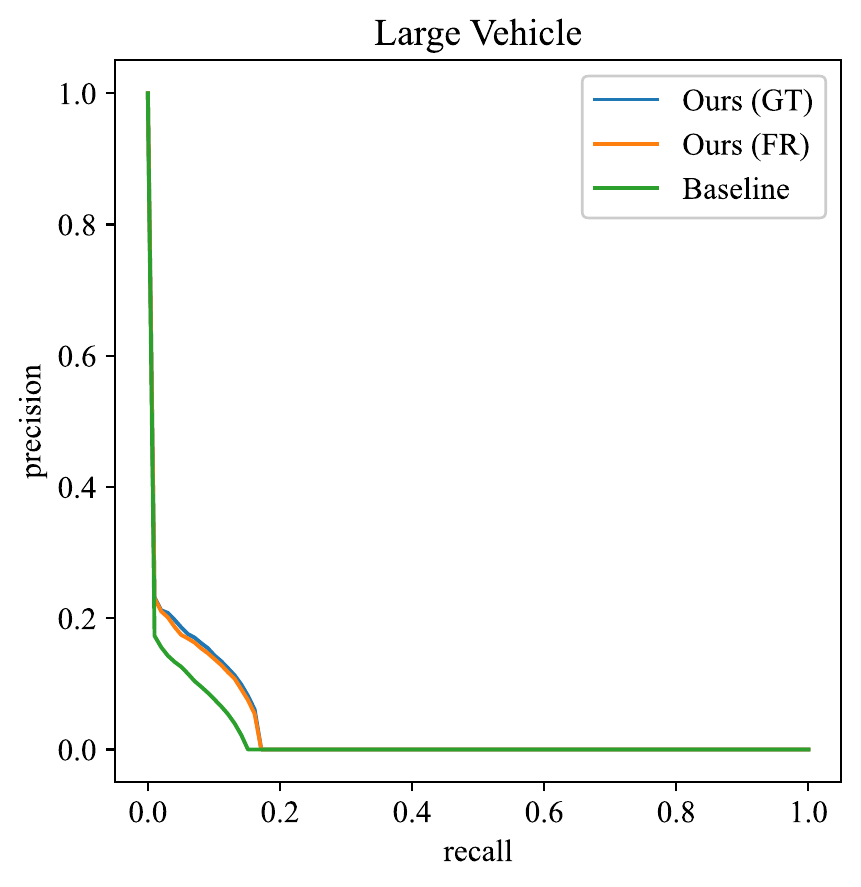}
   \includegraphics[width=0.19\linewidth]{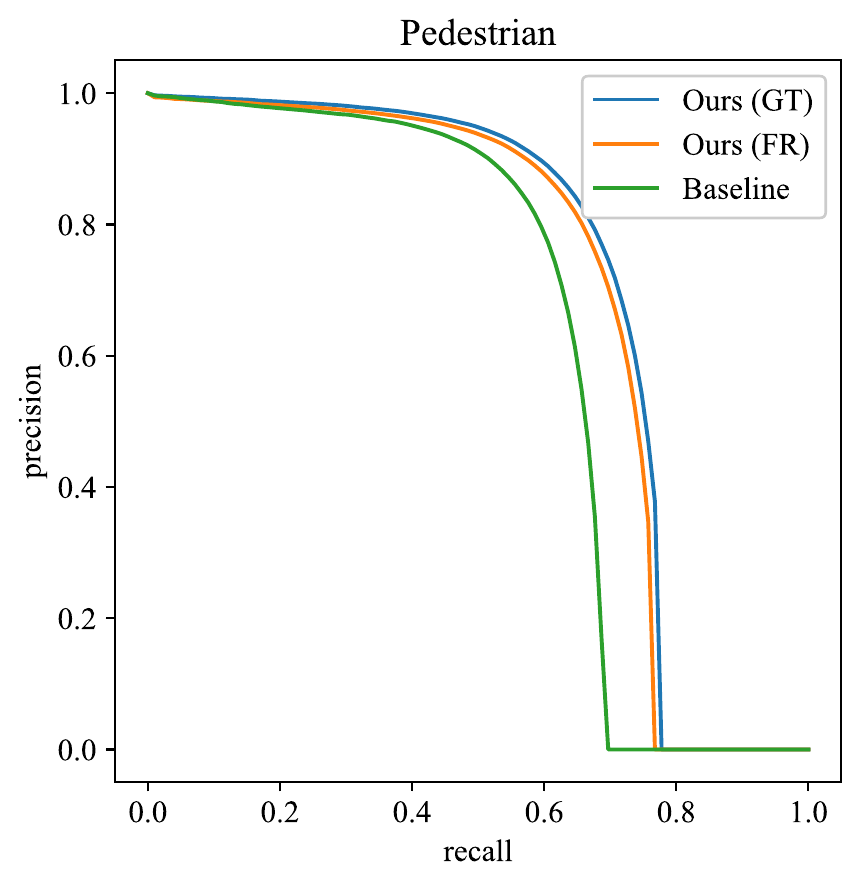}
   \includegraphics[width=0.19\linewidth]{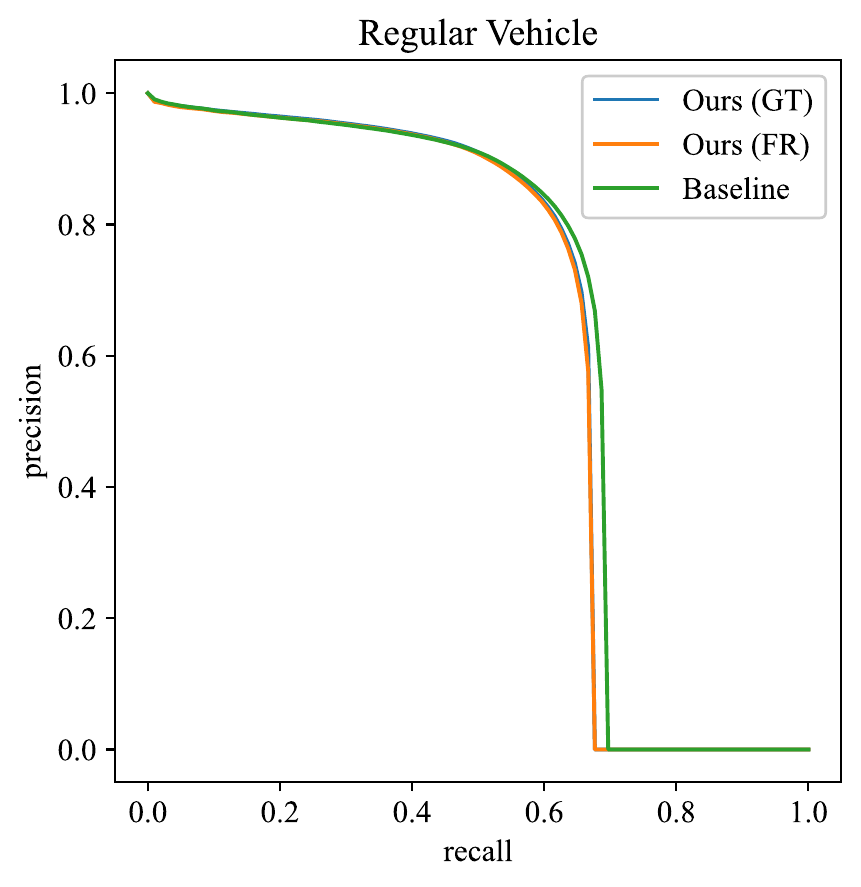}
   \includegraphics[width=0.19\linewidth]{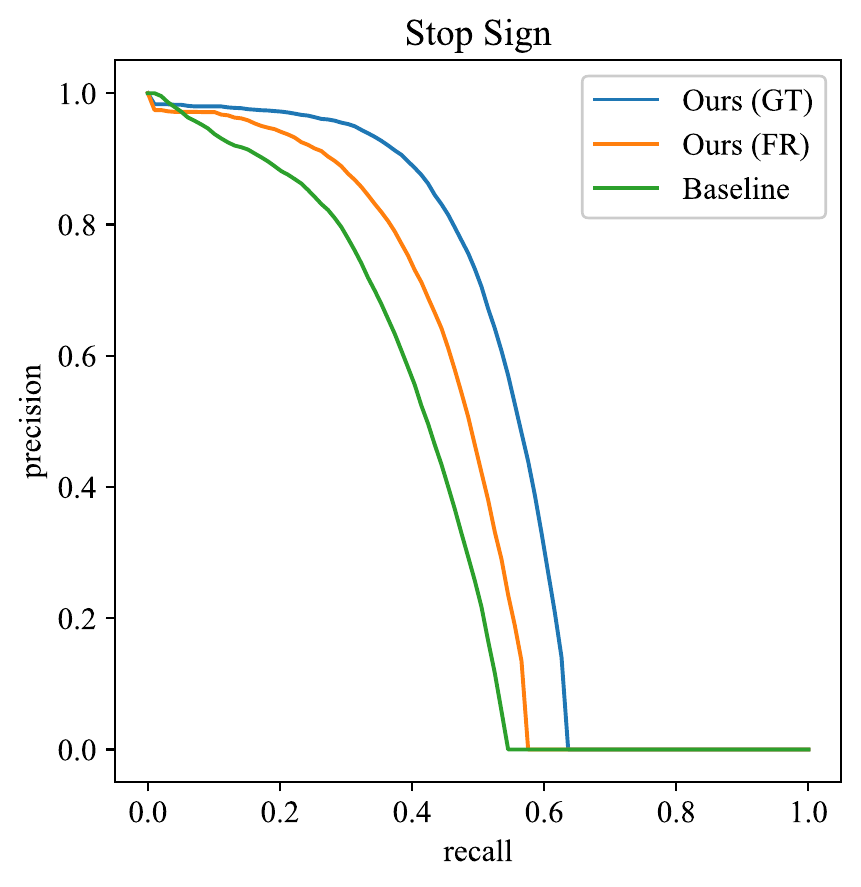}
   \includegraphics[width=0.19\linewidth]{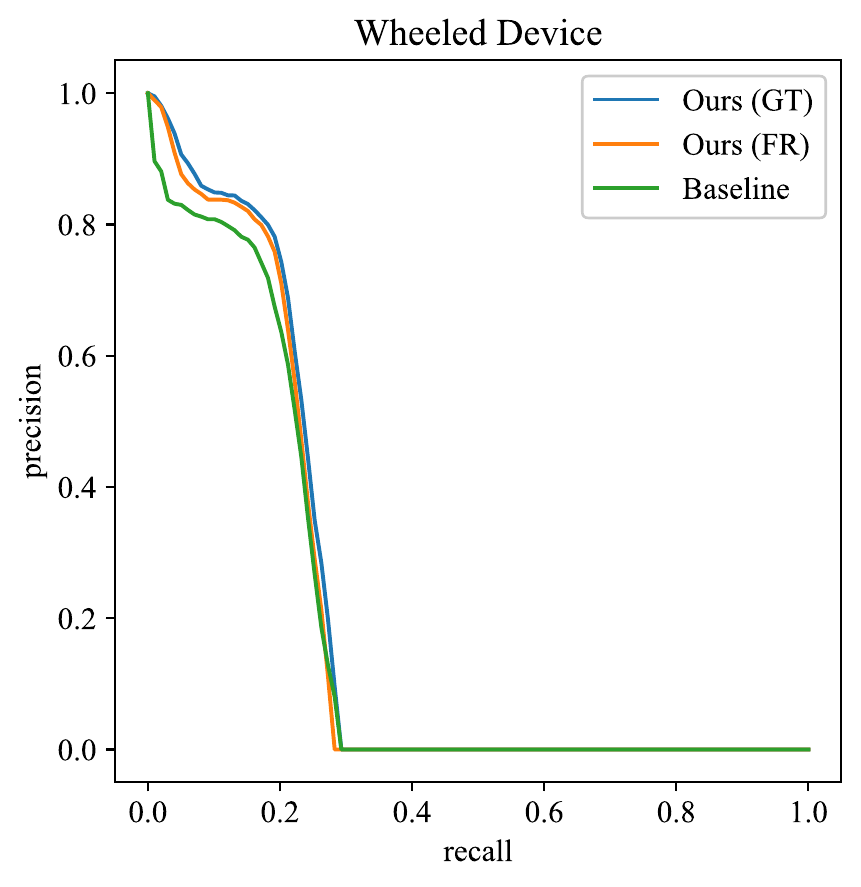}
   \caption{Precision-recall curves of the top-10 object classes with the most instance counts in the AV2 dataset. GT indicates we adopt the ground-truth bounding boxes as inputs. FR indicates we adopt bounding boxes generated by Faster R-CNN \cite{renNIPS15fasterrcnn}. }
   \label{fig:pr-curve}
\end{figure*}

This section provides more implementation details of our proposed method and experiments. All experiments were trained using mixed precision. Following previous methods \cite{liu2023bevfusion, bai2022transfusion}, we first train the LiDAR branch for 20 epochs with copy-and-paste augmentation strategy \cite{yin2021centerpoint} and disable this augmentation for the last 5 epochs. Then we train the LiDAR-camera fusion module for another 6 epochs. We adopt CBGS \cite{zhu2019CBGS} for class-balance learning. Experiments are trained with 16 RTX 3090 GPU. We decay the learning rate with a cosine annealing and employ AdamW \cite{loshchilov2017adamw} with a weight decay of 1e-2. 

For nuScenes \cite{caesar2020nuscenes} dataset, the learning rate is set to 4e-4, using batch size 4 on each GPU. For Argoverse2 \cite{Argoverse2} dataset, we use batch size 2 per GPU for our method, and batch size 1 per GPU for BEVFusion \cite{liang2022bevfusion}. For fair comparison and mitigating the impact of different batch size, we add SyncBN \cite{peng2018megdet} to BEVFusion on the Argoverse2 dataset. 

\section*{OpenLane}

\noindent\textbf{OpenLane Dataset. } OpenLane dataset is a large-scale 3D lane detection dataset built based on the Waymo \cite{sun2020waymo} dataset. It contains 1000 segments, 200K frames, and over 880K annotated lanes. It includes LiDAR and camera inputs and provides 2D and 3D lane annotations. The OpenLane dataset only has a forward camera with a perception range of 100m (cover area of 20m $\times$ 100m). We evaluated F1-score on the OpenLane dataset.

\noindent\textbf{3D Lane Detection.} Following BEV-LaneDet \cite{wang2022bevlanedet}, we set the BEV grid size to (0.5m, 0.5m). The perception range is set as [3m, 103m] for the X-axis, [-12m, 12m] for the Y-axis, and [-4m, 4m] for the Z-axis. For the image encoder, we use the same ResNet34 \cite{he2016resnet} as the image backbone and resize the image to 576$\times$1024. As for the LiDAR encoder, we use VoxelNet as the LiDAR backbone and set the voxel size to (0.125m, 0.125m, 0.2m). We adopt the 2D lane segmentation branch used in BEV-LaneDet to generate image masks. we train the LiDAR-camera fusion module for 10 epochs with 8 RTX 3090 GPU, and using batch size 8 per GPU. We set the learning rate to 2e-3. The other settings are similar to Argoverse2. 

\section*{Experimental results}

\noindent\textbf{BEV Sparsity.} We calculate the memory cost and latency of our method across various BEV sparsity under the Argoverse2 dataset, as depicted in Fig. \ref{fig:sparsity_latency_mem}. The results reveal a discernible trend: the memory cost and latency of SparseFusion exhibit a decline as BEV sparsity increases. Nevertheless, it's worth noting that owing to the inherent overhead of sparse convolution \cite{yan2018second}, SparseFusion's overhead surpasses that of BEVFusion when the BEV exhibits less than 47\% sparsity. To fully harness the benefits of sparse convolution for accelerating inference speed and minimizing memory cost, we find it imperative to maintain the BEV sparsity exceeding 70\%.

\noindent\textbf{2D box mask strategy.} During training, exclusively preserving regions corresponding to GT boxes results in the filtration of a significant portion of background regions, thereby diminishing the model's capability to accurately differentiate between foreground and background elements. To address this limitation, we adopt a strategy of randomly retaining a subset of background regions during training. Specifically, we introduce randomly initialized noise windows, which are utilized alongside the GT 2D boxes as foreground masks. During testing, only the predicted boxes are employed, mitigating any increase in memory and time overhead. The experiment result presented in Tab. \ref{tab:2d_box}, demonstrates an enhancement in model performance by 0.4 mAP following the incorporation of noise windows.

\begin{table}\footnotesize
    \centering
    \begin{tabular}{l|ccc|c}
        \toprule
                       & 0-50m & 50-100m & 100-200m & 0-200m \\
        \midrule
        SparseFusion-C & 0.316 & 0.129   & 0.043    & 0.221  \\
        SparseFusion-L & 0.464 & 0.230   & 0.078    & 0.329  \\
        SparseFusion   & \textbf{0.542} & \textbf{0.286}   & 0.105    & \textbf{0.398}  \\
        BEVFusion      & 0.528 & 0.284   & \textbf{0.113}    & 0.388 \\
        \bottomrule
    \end{tabular}
    \caption{The mAP of SparseFusion and BEVFusion under different perception ranges. C means the model takes the image as input. L means the model takes the point cloud as input. }
    \label{tab:range}
\end{table}

\begin{table}[]
    \centering
    \begin{tabular}{l|cc}
            \toprule
            & mAP & CDS  \\ 
            \midrule
            w/o. noise window & 0.394 & 0.307 \\ 
            w/. noise window  & 0.398 & 0.310 \\
            \bottomrule
        \end{tabular}
    \caption{Ablation of noise window on AV2 val set. }
    \label{tab:2d_box}
\end{table}

\noindent\textbf{Recall Analysis.} In Fig. \ref{fig:pr-curve}, we present the precision-recall curves for the top 10 object classes with the highest instance counts in the AV2 dataset. These curves illustrate that our model achieves higher precision and recall compared to the baseline, particularly for small objects such as pedestrians, bollards, and stop signs.

\noindent\textbf{Performance in various ranges.} We analyze SparseFusion's performance across various modality inputs and perception ranges. As detailed in Tab. \ref{tab:range}, our model reveals a significant enhancement in detection performance with multi-modal inputs compared to single-modal inputs. This improvement is attributed to the synergistic integration of LiDAR's geometric information and the camera's semantic information in our multi-modal fusion model. Furthermore, our model demonstrates superior performance over BEVFusion, with an enhanced mAP of 1.4 and 0.2 at distances of 0-50m and 50-100m, respectively, albeit showing a slight decrease of 0.8 at a distance of 100-200 meters. We present our model's sparsity across different ranges, as depicted in Fig. \ref{fig:range_sparsity}. Notably, only 1.4\% of the BEV grid contains non-empty features beyond 100m. This emphasizes the efficiency of our sparse processing approach, as our model achieves comparable results with minimal features.

%% file: main.bbl
\begin{thebibliography}{67}
\providecommand{\natexlab}[1]{#1}
\providecommand{\url}[1]{\texttt{#1}}
\expandafter\ifx\csname urlstyle\endcsname\relax
  \providecommand{\doi}[1]{doi: #1}\else
  \providecommand{\doi}{doi: \begingroup \urlstyle{rm}\Url}\fi

\bibitem[Bai et~al.(2022)Bai, Hu, Zhu, Huang, Chen, Fu, and Tai]{bai2022transfusion}
Xuyang Bai, Zeyu Hu, Xinge Zhu, Qingqiu Huang, Yilun Chen, Hongbo Fu, and Chiew-Lan Tai.
\newblock {TransFusion}: Robust lidar-camera fusion for 3d object detection with transformers.
\newblock In \emph{CVPR}, pages 1090--1099, 2022.

\bibitem[Caesar et~al.(2020)Caesar, Bankiti, Lang, Vora, Liong, Xu, Krishnan, Pan, Baldan, and Beijbom]{caesar2020nuscenes}
Holger Caesar, Varun Bankiti, Alex~H Lang, Sourabh Vora, Venice~Erin Liong, Qiang Xu, Anush Krishnan, Yu Pan, Giancarlo Baldan, and Oscar Beijbom.
\newblock {nuScenes}: A multimodal dataset for autonomous driving.
\newblock In \emph{CVPR}, pages 11621--11631, 2020.

\bibitem[Cai et~al.(2023)Cai, Zhang, Zhou, Li, Ding, and Zhao]{cai2023bevfusion4d}
Hongxiang Cai, Zeyuan Zhang, Zhenyu Zhou, Ziyin Li, Wenbo Ding, and Jiuhua Zhao.
\newblock Bevfusion4d: Learning lidar-camera fusion under bird's-eye-view via cross-modality guidance and temporal aggregation.
\newblock \emph{arXiv preprint arXiv:2303.17099}, 2023.

\bibitem[Chen et~al.(2022)Chen, Sima, Li, Zheng, Xu, Geng, Li, He, Shi, Qiao, and Yan]{chen2022persformer}
Li Chen, Chonghao Sima, Yang Li, Zehan Zheng, Jiajie Xu, Xiangwei Geng, Hongyang Li, Conghui He, Jianping Shi, Yu Qiao, and Junchi Yan.
\newblock {PersFormer}: 3d lane detection via perspective transformer and the openlane benchmark.
\newblock In \emph{ECCV}, 2022.

\bibitem[Chen et~al.(2023{\natexlab{a}})Chen, Zhang, Wang, Wang, and Zhao]{chen2023futr3d}
Xuanyao Chen, Tianyuan Zhang, Yue Wang, Yilun Wang, and Hang Zhao.
\newblock {FUTR3D}: A unified sensor fusion framework for 3d detection.
\newblock In \emph{CVPR}, pages 172--181, 2023{\natexlab{a}}.

\bibitem[Chen et~al.(2023{\natexlab{b}})Chen, Liu, Zhang, Qi, and Jia]{chen2023voxelnext}
Yukang Chen, Jianhui Liu, Xiangyu Zhang, Xiaojuan Qi, and Jiaya Jia.
\newblock {VoxelNeXt}: Fully sparse voxelnet for 3d object detection and tracking.
\newblock In \emph{CVPR}, 2023{\natexlab{b}}.

\bibitem[Fan et~al.(2022{\natexlab{a}})Fan, Pang, Zhang, Wang, Zhao, Wang, Wang, and Zhang]{fan2022sst}
Lue Fan, Ziqi Pang, Tianyuan Zhang, Yu-Xiong Wang, Hang Zhao, Feng Wang, Naiyan Wang, and Zhaoxiang Zhang.
\newblock {Embracing Single Stride 3D Object Detector with Sparse Transformer}.
\newblock In \emph{CVPR}, 2022{\natexlab{a}}.

\bibitem[Fan et~al.(2022{\natexlab{b}})Fan, Wang, Wang, and Zhang]{fan2022fsd}
Lue Fan, Feng Wang, Naiyan Wang, and Zhaoxiang Zhang.
\newblock {Fully Sparse 3D Object Detection}.
\newblock In \emph{NeurIPS}, 2022{\natexlab{b}}.

\bibitem[Fan et~al.(2023)Fan, Wang, Wang, and Zhang]{fan2023fsdv2}
Lue Fan, Feng Wang, Naiyan Wang, and Zhaoxiang Zhang.
\newblock {FSD V2}: Improving fully sparse 3d object detection with virtual voxels.
\newblock \emph{arXiv preprint arXiv:2308.03755}, 2023.

\bibitem[Garnett et~al.(2019)Garnett, Cohen, Pe'er, Lahav, and Levi]{garnett20193dlanenet}
Noa Garnett, Rafi Cohen, Tomer Pe'er, Roee Lahav, and Dan Levi.
\newblock {3D-LaneNet}: end-to-end 3d multiple lane detection.
\newblock In \emph{ICCV}, pages 2921--2930, 2019.

\bibitem[Ge et~al.(2023)Ge, Chen, Xie, Wang, Hong, Lu, Li, and Luo]{ge2023metabev}
Chongjian Ge, Junsong Chen, Enze Xie, Zhongdao Wang, Lanqing Hong, Huchuan Lu, Zhenguo Li, and Ping Luo.
\newblock {MetaBEV}: Solving sensor failures for 3d detection and map segmentation.
\newblock In \emph{ICCV}, pages 8721--8731, 2023.

\bibitem[Graham et~al.(2018)Graham, Engelcke, and Van Der~Maaten]{graham20183subm}
Benjamin Graham, Martin Engelcke, and Laurens Van Der~Maaten.
\newblock 3d semantic segmentation with submanifold sparse convolutional networks.
\newblock In \emph{CVPR}, pages 9224--9232, 2018.

\bibitem[Han et~al.(2023)Han, Sun, Ge, Yang, Dong, Zhou, Mao, Peng, and Zhang]{han2023videobev}
Chunrui Han, Jianjian Sun, Zheng Ge, Jinrong Yang, Runpei Dong, Hongyu Zhou, Weixin Mao, Yuang Peng, and Xiangyu Zhang.
\newblock Exploring recurrent long-term temporal fusion for multi-view 3d perception.
\newblock \emph{arXiv preprint arXiv:2303.05970}, 2023.

\bibitem[He et~al.(2016)He, Zhang, Ren, and Sun]{he2016resnet}
Kaiming He, Xiangyu Zhang, Shaoqing Ren, and Jian Sun.
\newblock Deep residual learning for image recognition.
\newblock In \emph{CVPR}, pages 770--778, 2016.

\bibitem[Hu et~al.(2023{\natexlab{a}})Hu, Zheng, Li, Xu, Mao, Luo, Wang, Chen, Liu, Zhao, et~al.]{hu2023fusionformer}
Chunyong Hu, Hang Zheng, Kun Li, Jianyun Xu, Weibo Mao, Maochun Luo, Lingxuan Wang, Mingxia Chen, Kaixuan Liu, Yiru Zhao, et~al.
\newblock {FusionFormer}: A multi-sensory fusion in bird's-eye-view and temporal consistent transformer for 3d objection.
\newblock \emph{arXiv preprint arXiv:2309.05257}, 2023{\natexlab{a}}.

\bibitem[Hu et~al.(2023{\natexlab{b}})Hu, Wang, Su, Wang, Hu, Fang, Xu, and Zhang]{hu2023ealss}
Haotian Hu, Fanyi Wang, Jingwen Su, Yaonong Wang, Laifeng Hu, Weiye Fang, Jingwei Xu, and Zhiwang Zhang.
\newblock {EA-LSS}: Edge-aware lift-splat-shot framework for 3d bev object detection.
\newblock \emph{arXiv preprint arXiv:2303.17895}, 2023{\natexlab{b}}.

\bibitem[Huang and Huang(2022{\natexlab{a}})]{huang2022bevdet4d}
Junjie Huang and Guan Huang.
\newblock Bevdet4d: Exploit temporal cues in multi-camera 3d object detection.
\newblock \emph{arXiv preprint arXiv:2203.17054}, 2022{\natexlab{a}}.

\bibitem[Huang and Huang(2022{\natexlab{b}})]{huang2022bevpoolv2}
Junjie Huang and Guan Huang.
\newblock {BEVPoolv2}: A cutting-edge implementation of bevdet toward deployment.
\newblock \emph{arXiv preprint arXiv:2211.17111}, 2022{\natexlab{b}}.

\bibitem[Huang et~al.(2021)Huang, Huang, Zhu, Yun, and Du]{huang2021bevdet}
Junjie Huang, Guan Huang, Zheng Zhu, Ye Yun, and Dalong Du.
\newblock {BEVDet}: High-performance multi-camera 3d object detection in bird-eye-view.
\newblock \emph{arXiv preprint arXiv:2112.11790}, 2021.

\bibitem[Huang et~al.(2023)Huang, Shen, Huang, Ding, Dai, Han, Wang, and Liu]{huang2023anchor3dlane}
Shaofei Huang, Zhenwei Shen, Zehao Huang, Zi-han Ding, Jiao Dai, Jizhong Han, Naiyan Wang, and Si Liu.
\newblock {Anchor3DLane}: Learning to regress 3d anchors for monocular 3d lane detection.
\newblock In \emph{CVPR}, 2023.

\bibitem[Jiang et~al.(2023)Jiang, Li, Liu, Wang, Jia, Wang, Han, and Zhang]{jiang2023far3d}
Xiaohui Jiang, Shuailin Li, Yingfei Liu, Shihao Wang, Fan Jia, Tiancai Wang, Lijin Han, and Xiangyu Zhang.
\newblock {Far3D}: Expanding the horizon for surround-view 3d object detection.
\newblock \emph{arXiv preprint arXiv:2308.09616}, 2023.

\bibitem[Lang et~al.(2019)Lang, Vora, Caesar, Zhou, Yang, and Beijbom]{lang2019pointpillars}
Alex~H Lang, Sourabh Vora, Holger Caesar, Lubing Zhou, Jiong Yang, and Oscar Beijbom.
\newblock {PointPillars}: Fast encoders for object detection from point clouds.
\newblock In \emph{CVPR}, pages 12697--12705, 2019.

\bibitem[Li et~al.(2022{\natexlab{a}})Li, Bao, Ge, Yang, Sun, and Li]{li2022bevstereo}
Yinhao Li, Han Bao, Zheng Ge, Jinrong Yang, Jianjian Sun, and Zeming Li.
\newblock {BEVStereo}: Enhancing depth estimation in multi-view 3d object detection with dynamic temporal stereo.
\newblock \emph{arXiv preprint arXiv:2209.10248}, 2022{\natexlab{a}}.

\bibitem[Li et~al.(2023{\natexlab{a}})Li, Fan, Liu, Huang, Chen, Wang, Zhang, and Tan]{li2023fsf}
Yingyan Li, Lue Fan, Yang Liu, Zehao Huang, Yuntao Chen, Naiyan Wang, Zhaoxiang Zhang, and Tieniu Tan.
\newblock Fully sparse fusion for 3d object detection.
\newblock \emph{arXiv preprint arXiv:2304.12310}, 2023{\natexlab{a}}.

\bibitem[Li et~al.(2023{\natexlab{b}})Li, Ge, Yu, Yang, Wang, Shi, Sun, and Li]{li2023bevdepth}
Yinhao Li, Zheng Ge, Guanyi Yu, Jinrong Yang, Zengran Wang, Yukang Shi, Jianjian Sun, and Zeming Li.
\newblock {BEVDepth}: Acquisition of reliable depth for multi-view 3d object detection.
\newblock In \emph{AAAI}, pages 1477--1485, 2023{\natexlab{b}}.

\bibitem[Li et~al.(2023{\natexlab{c}})Li, Huang, Chen, Cui, Liang, Shen, Liu, Xie, Sheng, Ouyang, et~al.]{li2023fastbev}
Yangguang Li, Bin Huang, Zeren Chen, Yufeng Cui, Feng Liang, Mingzhu Shen, Fenggang Liu, Enze Xie, Lu Sheng, Wanli Ouyang, et~al.
\newblock {Fast-BEV}: A fast and strong bird's-eye view perception baseline.
\newblock \emph{arXiv preprint arXiv:2301.12511}, 2023{\natexlab{c}}.

\bibitem[Li et~al.(2022{\natexlab{b}})Li, Wang, Li, Xie, Sima, Lu, Qiao, and Dai]{li2022bevformer}
Zhiqi Li, Wenhai Wang, Hongyang Li, Enze Xie, Chonghao Sima, Tong Lu, Yu Qiao, and Jifeng Dai.
\newblock {BEVFormer}: Learning bird’s-eye-view representation from multi-camera images via spatiotemporal transformers.
\newblock In \emph{ECCV}, pages 1--18. Springer, 2022{\natexlab{b}}.

\bibitem[Liang et~al.(2022)Liang, Xie, Yu, Xia, Lin, Wang, Tang, Wang, and Tang]{liang2022bevfusion}
Tingting Liang, Hongwei Xie, Kaicheng Yu, Zhongyu Xia, Zhiwei Lin, Yongtao Wang, Tao Tang, Bing Wang, and Zhi Tang.
\newblock {BEVFusion}: A simple and robust lidar-camera fusion framework.
\newblock \emph{NeurIPS}, 35:\penalty0 10421--10434, 2022.

\bibitem[Lin et~al.(2017)Lin, Doll{\'a}r, Girshick, He, Hariharan, and Belongie]{lin2017fpn}
Tsung-Yi Lin, Piotr Doll{\'a}r, Ross Girshick, Kaiming He, Bharath Hariharan, and Serge Belongie.
\newblock Feature pyramid networks for object detection.
\newblock In \emph{CVPR}, pages 2117--2125, 2017.

\bibitem[Lin et~al.(2022)Lin, Lin, Pei, Huang, and Su]{lin2022sparse4d}
Xuewu Lin, Tianwei Lin, Zixiang Pei, Lichao Huang, and Zhizhong Su.
\newblock {Sparse4D}: Multi-view 3d object detection with sparse spatial-temporal fusion.
\newblock \emph{arXiv preprint arXiv:2211.10581}, 2022.

\bibitem[Liu et~al.(2023{\natexlab{a}})Liu, Teng, Lu, Wang, and Wang]{liu2023sparsebev}
Haisong Liu, Yao Teng, Tao Lu, Haiguang Wang, and Limin Wang.
\newblock {SparseBEV}: High-performance sparse 3d object detection from multi-camera videos.
\newblock In \emph{ICCV}, pages 18580--18590, 2023{\natexlab{a}}.

\bibitem[Liu et~al.(2022{\natexlab{a}})Liu, Wang, Zhang, and Sun]{liu2022petr}
Yingfei Liu, Tiancai Wang, Xiangyu Zhang, and Jian Sun.
\newblock {PETR}: Position embedding transformation for multi-view 3d object detection.
\newblock In \emph{ECCV}, pages 531--548. Springer, 2022{\natexlab{a}}.

\bibitem[Liu et~al.(2022{\natexlab{b}})Liu, Yan, Jia, Li, Gao, Wang, Zhang, and Sun]{liu2022petrv2}
Yingfei Liu, Junjie Yan, Fan Jia, Shuailin Li, Aqi Gao, Tiancai Wang, Xiangyu Zhang, and Jian Sun.
\newblock {PETRv2}: A unified framework for 3d perception from multi-camera images.
\newblock \emph{arXiv preprint arXiv:2206.01256}, 2022{\natexlab{b}}.

\bibitem[Liu et~al.(2021)Liu, Lin, Cao, Hu, Wei, Zhang, Lin, and Guo]{liu2021swin}
Ze Liu, Yutong Lin, Yue Cao, Han Hu, Yixuan Wei, Zheng Zhang, Stephen Lin, and Baining Guo.
\newblock {Swin Transformer}: Hierarchical vision transformer using shifted windows.
\newblock In \emph{CVPR}, pages 10012--10022, 2021.

\bibitem[Liu et~al.(2023{\natexlab{b}})Liu, Tang, Amini, Yang, Mao, Rus, and Han]{liu2023bevfusion}
Zhijian Liu, Haotian Tang, Alexander Amini, Xinyu Yang, Huizi Mao, Daniela~L Rus, and Song Han.
\newblock {BEVFusion}: Multi-task multi-sensor fusion with unified bird's-eye view representation.
\newblock In \emph{ICRA}, pages 2774--2781. IEEE, 2023{\natexlab{b}}.

\bibitem[Liu et~al.(2023{\natexlab{c}})Liu, Yang, Tang, Yang, and Han]{liu2023flatformer}
Zhijian Liu, Xinyu Yang, Haotian Tang, Shang Yang, and Song Han.
\newblock {FlatFormer}: Flattened window attention for efficient point cloud transformer.
\newblock In \emph{CVPR}, pages 1200--1211, 2023{\natexlab{c}}.

\bibitem[Loshchilov and Hutter(2017)]{loshchilov2017adamw}
Ilya Loshchilov and Frank Hutter.
\newblock Decoupled weight decay regularization.
\newblock \emph{arXiv preprint arXiv:1711.05101}, 2017.

\bibitem[{Luo} et~al.(2022){Luo}, {Yan}, {Zheng}, {Zheng}, {Mei}, {Kun}, {Cui}, and {Li}]{2022arXivm2-3dlane}
Yueru {Luo}, Xu {Yan}, Chaoda {Zheng}, Chao {Zheng}, Shuqi {Mei}, Tang {Kun}, Shuguang {Cui}, and Zhen {Li}.
\newblock {M$^2$-3DLaneNet: Exploring Multi-Modal 3D Lane Detection}.
\newblock \emph{arXiv e-prints}, art. arXiv:2209.05996, 2022.

\bibitem[Luo et~al.(2023)Luo, Zheng, Yan, Kun, Zheng, Cui, and Li]{luo2023latr}
Yueru Luo, Chaoda Zheng, Xu Yan, Tang Kun, Chao Zheng, Shuguang Cui, and Zhen Li.
\newblock {LATR}: 3d lane detection from monocular images with transformer.
\newblock In \emph{ICCV}, pages 7941--7952, 2023.

\bibitem[Pei et~al.(2023)Pei, Zhao, Li, Ma, Zhang, and Pu]{pei2023clusterformer}
Yu Pei, Xian Zhao, Hao Li, Jingyuan Ma, Jingwei Zhang, and Shiliang Pu.
\newblock {Cluster-Former}: Cluster-based transformer for 3d object detection in point clouds.
\newblock In \emph{ICCV}, pages 6664--6673, 2023.

\bibitem[Peng et~al.(2018)Peng, Xiao, Li, Jiang, Zhang, Jia, Yu, and Sun]{peng2018megdet}
Chao Peng, Tete Xiao, Zeming Li, Yuning Jiang, Xiangyu Zhang, Kai Jia, Gang Yu, and Jian Sun.
\newblock Megdet: A large mini-batch object detector.
\newblock In \emph{Proceedings of the IEEE conference on Computer Vision and Pattern Recognition}, pages 6181--6189, 2018.

\bibitem[Philion and Fidler(2020)]{philion2020lift}
Jonah Philion and Sanja Fidler.
\newblock Lift, splat, shoot: Encoding images from arbitrary camera rigs by implicitly unprojecting to 3d.
\newblock In \emph{ECCV}, pages 194--210. Springer, 2020.

\bibitem[Qi et~al.(2017)Qi, Su, Mo, and Guibas]{qi2017pointnet}
Charles~R Qi, Hao Su, Kaichun Mo, and Leonidas~J Guibas.
\newblock {PointNet}: Deep learning on point sets for 3d classification and segmentation.
\newblock In \emph{CVPR}, pages 652--660, 2017.

\bibitem[Redmon et~al.(2016)Redmon, Divvala, Girshick, and Farhadi]{redmon2016yolo}
Joseph Redmon, Santosh Divvala, Ross Girshick, and Ali Farhadi.
\newblock {You Only Look Once}: Unified, real-time object detection.
\newblock In \emph{CVPR}, pages 779--788, 2016.

\bibitem[Ren et~al.(2015)Ren, He, Girshick, and Sun]{renNIPS15fasterrcnn}
Shaoqing Ren, Kaiming He, Ross Girshick, and Jian Sun.
\newblock Faster {R-CNN}: Towards real-time object detection with region proposal networks.
\newblock In \emph{NeurIPS}, 2015.

\bibitem[Sun et~al.(2020)Sun, Kretzschmar, Dotiwalla, Chouard, Patnaik, Tsui, Guo, Zhou, Chai, Caine, et~al.]{sun2020waymo}
Pei Sun, Henrik Kretzschmar, Xerxes Dotiwalla, Aurelien Chouard, Vijaysai Patnaik, Paul Tsui, James Guo, Yin Zhou, Yuning Chai, Benjamin Caine, et~al.
\newblock Scalability in perception for autonomous driving: Waymo open dataset.
\newblock In \emph{CVPR}, pages 2446--2454, 2020.

\bibitem[Tian et~al.(2019)Tian, Shen, Chen, and He]{tian2019fcos}
Zhi Tian, Chunhua Shen, Hao Chen, and Tong He.
\newblock {FCOS}: Fully convolutional one-stage object detection.
\newblock In \emph{ICCV}, 2019.

\bibitem[Vaswani et~al.(2017)Vaswani, Shazeer, Parmar, Uszkoreit, Jones, Gomez, Kaiser, and Polosukhin]{vaswani2017attention}
Ashish Vaswani, Noam Shazeer, Niki Parmar, Jakob Uszkoreit, Llion Jones, Aidan~N Gomez, {\L}ukasz Kaiser, and Illia Polosukhin.
\newblock Attention is all you need.
\newblock \emph{NeurIPS}, 30, 2017.

\bibitem[Wang et~al.(2023{\natexlab{a}})Wang, Shi, Shi, Lei, Wang, He, Schiele, and Wang]{wang2023dsvt}
Haiyang Wang, Chen Shi, Shaoshuai Shi, Meng Lei, Sen Wang, Di He, Bernt Schiele, and Liwei Wang.
\newblock {DSVT}: Dynamic sparse voxel transformer with rotated sets.
\newblock In \emph{CVPR}, pages 13520--13529, 2023{\natexlab{a}}.

\bibitem[Wang et~al.(2023{\natexlab{b}})Wang, Tang, Shi, Li, Li, Schiele, and Wang]{wang2023unitr}
Haiyang Wang, Hao Tang, Shaoshuai Shi, Aoxue Li, Zhenguo Li, Bernt Schiele, and Liwei Wang.
\newblock {UniTR}: A unified and efficient multi-modal transformer for bird's-eye-view representation.
\newblock In \emph{ICCV}, pages 6792--6802, 2023{\natexlab{b}}.

\bibitem[Wang et~al.(2022)Wang, Qin, Li, and Cao]{wang2022bevlanedet}
Ruihao Wang, Jian Qin, Kaiying Li, and Dong Cao.
\newblock {BEV-LaneDet}: Fast lane detection on bev ground.
\newblock \emph{arXiv preprint arXiv:2210.06006}, 2022.

\bibitem[Wang et~al.(2023{\natexlab{c}})Wang, Caesar, Nan, and Kooij]{wang2023unibev}
Shiming Wang, Holger Caesar, Liangliang Nan, and Julian~FP Kooij.
\newblock {UniBEV}: Multi-modal 3d object detection with uniform bev encoders for robustness against missing sensor modalities.
\newblock \emph{arXiv preprint arXiv:2309.14516}, 2023{\natexlab{c}}.

\bibitem[Wang et~al.(2023{\natexlab{d}})Wang, Liu, Wang, Li, and Zhang]{wang2023streampetr}
Shihao Wang, Yingfei Liu, Tiancai Wang, Ying Li, and Xiangyu Zhang.
\newblock Exploring object-centric temporal modeling for efficient multi-view 3d object detection.
\newblock \emph{arXiv preprint arXiv:2303.11926}, 2023{\natexlab{d}}.

\bibitem[Wang et~al.(2021)Wang, Guizilini, Zhang, Wang, Zhao, , and Solomon]{detr3d}
Yue Wang, Vitor Guizilini, Tianyuan Zhang, Yilun Wang, Hang Zhao, , and Justin~M. Solomon.
\newblock {DETR3D}: 3d object detection from multi-view images via 3d-to-2d queries.
\newblock In \emph{CoRL}, 2021.

\bibitem[Wang et~al.(2023{\natexlab{e}})Wang, Huang, Fu, Wang, and Liu]{wang2023mv2d}
Zitian Wang, Zehao Huang, Jiahui Fu, Naiyan Wang, and Si Liu.
\newblock Object as query: Equipping any 2d object detector with 3d detection ability.
\newblock \emph{arXiv preprint arXiv:2301.02364}, 2023{\natexlab{e}}.

\bibitem[Wilson et~al.(2021)Wilson, Qi, Agarwal, Lambert, Singh, Khandelwal, Pan, Kumar, Hartnett, Pontes, Ramanan, Carr, and Hays]{Argoverse2}
Benjamin Wilson, William Qi, Tanmay Agarwal, John Lambert, Jagjeet Singh, Siddhesh Khandelwal, Bowen Pan, Ratnesh Kumar, Andrew Hartnett, Jhony~Kaesemodel Pontes, Deva Ramanan, Peter Carr, and James Hays.
\newblock {Argoverse 2}: Next generation datasets for self-driving perception and forecasting.
\newblock In \emph{NeurIPS}, 2021.

\bibitem[Xie et~al.(2022)Xie, Yu, Zhou, Philion, Anandkumar, Fidler, Luo, and Alvarez]{xie2022m2bev}
Enze Xie, Zhiding Yu, Daquan Zhou, Jonah Philion, Anima Anandkumar, Sanja Fidler, Ping Luo, and Jose~M Alvarez.
\newblock M2bev: Multi-camera joint 3d detection and segmentation with unified birds-eye view representation.
\newblock \emph{arXiv preprint arXiv:2204.05088}, 2022.

\bibitem[Yan et~al.(2023)Yan, Liu, Sun, Jia, Li, Wang, and Zhang]{yan2023cmt}
Junjie Yan, Yingfei Liu, Jianjian Sun, Fan Jia, Shuailin Li, Tiancai Wang, and Xiangyu Zhang.
\newblock {Cross Modal Transformer}: Towards fast and robust 3d object detection.
\newblock In \emph{ICCV}, pages 18268--18278, 2023.

\bibitem[Yan et~al.(2018)Yan, Mao, and Li]{yan2018second}
Yan Yan, Yuxing Mao, and Bo Li.
\newblock {SECOND}: Sparsely embedded convolutional detection.
\newblock \emph{Sensors}, 18\penalty0 (10):\penalty0 3337, 2018.

\bibitem[Yang et~al.(2023)Yang, Chen, Tian, Tao, Zhu, Zhang, Huang, Li, Qiao, Lu, et~al.]{yang2023bevformerv2}
Chenyu Yang, Yuntao Chen, Hao Tian, Chenxin Tao, Xizhou Zhu, Zhaoxiang Zhang, Gao Huang, Hongyang Li, Yu Qiao, Lewei Lu, et~al.
\newblock {BEVFormer v2}: Adapting modern image backbones to bird's-eye-view recognition via perspective supervision.
\newblock In \emph{CVPR}, pages 17830--17839, 2023.

\bibitem[Yao et~al.(2023)Yao, Yu, Wu, and Jia]{yao2023sparsepoint}
Chengtang Yao, Lidong Yu, Yuwei Wu, and Yunde Jia.
\newblock Sparse point guided 3d lane detection.
\newblock In \emph{ICCV}, pages 8363--8372, 2023.

\bibitem[Yin et~al.(2021)Yin, Zhou, and Krahenbuhl]{yin2021centerpoint}
Tianwei Yin, Xingyi Zhou, and Philipp Krahenbuhl.
\newblock Center-based 3d object detection and tracking.
\newblock In \emph{CVPR}, pages 11784--11793, 2021.

\bibitem[Zhang et~al.(2023)Zhang, Zhang, Liu, and Wang]{zhang2023sabev}
Jinqing Zhang, Yanan Zhang, Qingjie Liu, and Yunhong Wang.
\newblock {SA-BEV}: Generating semantic-aware bird's-eye-view feature for multi-view 3d object detection.
\newblock In \emph{ICCV}, pages 3348--3357, 2023.

\bibitem[Zhou and Tuzel(2018)]{zhou2018voxelnet}
Yin Zhou and Oncel Tuzel.
\newblock {VoxelNet}: End-to-end learning for point cloud based 3d object detection.
\newblock In \emph{CVPR}, pages 4490--4499, 2018.

\bibitem[Zhou et~al.(2020)Zhou, Sun, Zhang, Anguelov, Gao, Ouyang, Guo, Ngiam, and Vasudevan]{zhou2020MVF}
Yin Zhou, Pei Sun, Yu Zhang, Dragomir Anguelov, Jiyang Gao, Tom Ouyang, James Guo, Jiquan Ngiam, and Vijay Vasudevan.
\newblock End-to-end multi-view fusion for 3d object detection in lidar point clouds.
\newblock In \emph{CoRL}, pages 923--932. PMLR, 2020.

\bibitem[Zhu et~al.(2019)Zhu, Jiang, Zhou, Li, and Yu]{zhu2019CBGS}
Benjin Zhu, Zhengkai Jiang, Xiangxin Zhou, Zeming Li, and Gang Yu.
\newblock Class-balanced grouping and sampling for point cloud 3d object detection.
\newblock \emph{arXiv preprint arXiv:1908.09492}, 2019.

\bibitem[Zhu et~al.(2020)Zhu, Su, Lu, Li, Wang, and Dai]{zhu2020deformable}
Xizhou Zhu, Weijie Su, Lewei Lu, Bin Li, Xiaogang Wang, and Jifeng Dai.
\newblock {Deformable DETR}: Deformable transformers for end-to-end object detection.
\newblock \emph{arXiv preprint arXiv:2010.04159}, 2020.

\end{thebibliography}
